%% file: main.tex
\documentclass{article}
\pdfoutput=1
\usepackage{arxiv}
\setcitestyle{authoryear,open={(},close={)}}
\renewcommand{\cite}{\citep}

\usepackage[utf8]{inputenc} %
\usepackage[T1]{fontenc}    %
\usepackage{booktabs}       
\usepackage{xcolor}
\usepackage{amsfonts}       
\usepackage{nicefrac}       
\usepackage{microtype}      
\usepackage{dsfont}
\usepackage{graphicx}
\usepackage{amsmath}
\usepackage{mathtools}
\usepackage{amssymb}
\usepackage{amsthm}
\usepackage{cases}
\usepackage{color}
\usepackage{textcomp}
\usepackage{caption}
\usepackage{bbm}
\usepackage{comment}
\usepackage{xspace}
\usepackage{multirow}
\usepackage{subcaption}
\usepackage{authblk}
\usepackage{xspace}
\usepackage{algorithm}
\usepackage{algpseudocode}
\usepackage{adjustbox}
\usepackage{wrapfig}
\usepackage{soul}
\usepackage{comment}

\usepackage{multirow}
\usepackage{setspace}
\usepackage{makecell}
\usepackage{graphicx}
\usepackage{parskip}
\usepackage{xcolor,colortbl}
\usepackage{soul}
\usepackage{color}
\usepackage{subcaption}
\usepackage{CJKutf8}
\usepackage{tcolorbox}\tcbuselibrary{skins}

\def\shownotes{0}
\ifnum\shownotes=1
\newcommand{\authnote}[2]{[#1: #2]}
\else
\newcommand{\authnote}[2]{}
\fi

\definecolor{pistachio}{rgb}{0.58, 0.77, 0.45}
\definecolor{asparagus}{rgb}{0.53, 0.66, 0.42}
\definecolor{cadmiumgreen}{rgb}{0.0, 0.42, 0.24}
\definecolor{cardinal}{rgb}{0.77, 0.12, 0.23}

\newcommand\yellowcell{\cellcolor[rgb]{1,0.843,0}}
\newcommand\faintedbluecell{\cellcolor[rgb]{0.85, 0.85, 1}}

\newcommand\blfootnote[1]{%
  \begingroup
  \renewcommand\thefootnote{}\footnote{#1}%
  \addtocounter{footnote}{-1}%
  \endgroup
}

\input{macros}

\input{std_macros}

\begin{document}
\bibliographystyle{plainnat}

\title{Educator Attention: \\ How computational tools can systematically identify the distribution of a key resource for students}

\author[]{
    \textbf{Qingyang Zhang}$^*$, 
    \textbf{Rose E. Wang}$^*$ \\
    \vspace{0.3em}
    \textbf{Ana T. Ribeiro},
    \textbf{Dora Demszky}$^\alpha$, 
    \textbf{Susanna Loeb}$^\alpha$
}

\affil[]{Stanford University \\ 
\textit{q1zhang@stanford, rewang@cs.stanford.edu}
}

\date{}

\newcommand{\fix}{\marginpar{FIX}}
\newcommand{\new}{\marginpar{NEW}}

\maketitle

\begin{abstract}
Educator attention is critical for student success, yet how educators distribute their attention across students remains poorly understood due to data and methodological constraints. 
This study presents the \textbf{first large-scale computational analysis of educator attention patterns}, leveraging over 1 million educator utterances from virtual group tutoring sessions linked to detailed student demographic and academic achievement data. 
Using natural language processing techniques, we systematically examine the recipient and nature of educator attention.
Our findings reveal that educators often provide more attention to lower-achieving students. 
However, disparities emerge across demographic lines, particularly by gender. 
Girls tend to receive less attention when paired with boys, even when they are the lower achieving student in the group. 
Lower-achieving female students in mixed-gender pairs receive significantly less attention than their higher-achieving male peers, while lower-achieving male students receive significantly and substantially more attention than their higher-achieving female peers. 
We also find some differences by race and English learner (EL) status, with low-achieving Black students receiving additional attention only when paired with another Black student but not when paired with a non-Black peer. 
In contrast, higher-achieving EL students receive disproportionately more attention than their lower-achieving EL peers. 
This work highlights how large-scale interaction data and computational methods can uncover subtle but meaningful disparities in teaching practices, providing empirical insights to inform more equitable and effective educational strategies.
\blfootnote{\hfill $* =$ Equal contributions. $\alpha = $ Equal advising.}

\end{abstract}

\section{Introduction}

Educator attention is one of the most valuable resources for building effective learning experiences---whether it's in the classroom, extra-curricular activities, or tutoring. 
When students receive more attention from educators, such as instructional or emotional support, students are more likely to learn and acquire the skills they need to engage successfully in school~\citep{hamre2005can, birch1997teacher, burchinal2008predicting, nickow2024promise, cortes2024scalable}. 
Recognizing the important role of educator's attention to student-specific needs, post-pandemic education policies have prioritized increasing the amount of attention students receive, such as by expanding tutoring programs \citep{groom2023challenges, future_ed}.

However, the presence of an educator does not guarantee that all students receive the same kind of attention.
Prior work has documented differences in how educators allocate attention, with student demographics influencing the quantity and nature of the attention students receive. 
Female students and students from racial minority groups, for instance, often receive less attention or lower quality instruction, such as being given easier content~\citep{beaman2006differential, mckown2008teacher, watson2022gender}. 
Educator attention is therefore a double-edged sword: While it has the potential to remediate education inequalities, it can also inadvertently perpetuate them. 

Understanding the patterns in educator attention requires large-scale, systematic analysis of both whom educators address and how they engage with students. 
Traditionally, researchers have relied on classroom observations where trained annotators manually track educator-student interactions, such as how often an educator calls on students~\citep{younger1999gender, allen2013observations, brown2010improving, gunn2021measuring}. 
While valuable, these observation methods are limited in scale, capturing only some hours of instruction, and lack systematic control over a student's academic background that shape educator attention beyond a student's demographics~\citep{reinholz2018equity}. 
For example, if educators interact more frequently with male students, is this due to gender alone or due to male students struggling more academically than female  students?

The recent expansion of virtual tutoring, where educators meet their students over video, has created unprecedented opportunities to study educator-student interactions at scale~\citep{gortazar2024online, cortes2025scalable}. 
Each interaction between educators and their students is recorded and linked with individual student characteristics. 
Coupled with advances in natural language processing (NLP), this data allows researchers to systematically analyze both the amount and nature of educator attention. 
Researchers have used NLP across domains to examine large-scale language patterns in high-stakes interactions, including police interactions~\citep{voigt2017language}, classroom interactions~\citep{jacobs2022promoting, suresh2022talkmoves} and tutoring interactions~\citep{demszky2024can, wang2024tutor}.  

In this work, we analyze over\numUtterances educator utterances in tutoring sessions and investigate how educator attention varies by student achievement and demographics. 
We focus on three key demographic indicators studied in prior work on educator attention: gender, race and English Learner (EL) status.
Our data comes from a randomized controlled trial of a tutoring program found to be effective at improving student literacy skills \citep{robinson2024effects}. 
Using NLP techniques, we automatically identify who educators direct their attention to and how they engage with the students. 

Our analysis provides the first large-scale quantitative study of educator attention patterns, where we both validate and extend previous qualitative research.
Concretely, we find that educators provide more attention to lower-achieving students, suggesting a responsiveness to student needs. 
However, we also reveal notable deviations from this pattern:
Lower-achieving female students in mixed-gender pairs receive significantly less attention than their higher-achieving male peers, whereas lower-achieving male students receive substantially more attention than their higher-achieving female peers. 
A similar pattern arises for Black students: Low-achieving Black students receive more attention than their peer only when paired with another Black student but not when paired with a non-Black peer. 
By contrast, higher-achieving EL students receive significantly more attention than their lower-achieving EL peers. 
These findings suggest that even within effective tutoring programs, student demographics and group compositions may unconsciously shape educator attention. 
Our study demonstrates how large-scale analysis of fine-grained educator-student interactions can uncover subtle but meaningful disparities, and surface opportunities to promote better teaching practices.

\input{images/framework}

\section*{Data}
Our dataset consists of transcribed video recordings from a U.S.-based virtual tutoring program focused on early literacy for K-2 students, which was found to improve student achievement scores in a randomized controlled trial \citep{robinson2024effects}.
This dataset includes \numUtterances utterances from \numTranscripts 2-on-1 tutoring sessions, where a tutor teaches two students in a session. 
Students in each pair were matched at the beginning of the school year, and largely remained in the same pair throughout the year for consistency.
The tutors are based in the U.S., and are mostly former classroom teachers and part-time teachers. 
Our dataset includes \numTeachers educators and \numStudents students. 

Every tutoring session is linked to student information provided by the school district, including gender, race, EL status, and achievement.
Given prior work on educator attention, we focus on three key demographic factors frequently studied in this literature, gender~\citep{beaman2006differential}, race~\citep{reinholz2018equity}, and EL status~\citep{solano2024equally}.
We also account for students’ pre-intervention achievement measured by their Dynamic Indicators of Basic Literacy Skills (DIBELS) scores, a common measurement of student literacy skills~\citep{good2001using}. 
Please refer to Appendix Section~\ref{app:data} for more details on data.

\section*{Measuring Attention}
Traditional classroom observations are effective in discovering patterns in educator attention, however are too resource-intensive for large-scale analysis.
Scalable, systematic approaches are important for advancing research on how educator attention varies in different settings and where to improve teaching practices.

In this study, we develop computational classification models of the recipient and of the nature of attention, and tune them on\numAnnotations~labeled utterances with preceding context from the interaction.
Our classification taxonomy, shown in Table~\ref{tab:framework}, synthesizes theoretical frameworks from prior education research on instructional attention with empirical observations from our dataset. 
This approach ensures our measures are both theoretically grounded and practically relevant.
See Materials and Methods for more detail.

By applying these trained models to our complete dataset of\numUtterances utterances, we conduct a systematic inquiry of educator attention patterns and their relationship to student characteristics and academic needs. 
Our work focuses on the educator's utterances, and we do not model student-initiated behaviors that can also shape how educators allocate attention~\citep{irvine1986teacher, brophy1970teachers}. 
Our findings characterize the distribution of educator attention and do not establish causal relationships.

\subsection*{Recipient of Attention}
Previous studies have documented differences in educator attention across students. 
For example, research has shown that classroom teachers give less individual attention to female students compared to male students~\citep{heller1981sex, jones2004meta, beaman2006differential}.
Studies of race differences have produced mixed results and research focusing on differential attention by EL status is sparse~\citep{cooper1998meta,irvine1986teacher, solano2024equally}.
Our computational approach provides the opportunity for us to explore differences across gender, race and EL status in educator attention by identifying the recipient of that attention leveraging past work on addressee identification~\citep{ouchi2016addressee, jovanovic2004towards}. 
In our two-student tutoring context, we classify each educator utterance into four categories: directed to student A, directed to student B, directed to both students, or directed to an ambiguous recipient. 
With this classification, we can quantify the amount of attention given to a student by measuring the total duration of utterances directed to that student. 
Additionally, by computationally measuring the amount of attention to a student, we can analyze how educator attention varies across student characteristics at a scale previously impossible with manual methods. 
Our main study focuses on the attention directed to individual students, and Appendix Section~\ref{app:add_results} reports results on attention directed to both students.

\subsection*{Nature of Attention}
Beyond measuring who receives attention, prior work has studied the differences in the nature of attention.
For example, researchers have found that boys receive more attention about their behaviors than girls do~\citep{dweck1978sex, irvine1986teacher}.
Prior work has also documented the importance of relationship-building and content-focused attention, because these forms of attention can help build student confidence and improve achievement especially for historically under-performing student groups including EL students~\citep{lewis2012carino, soland2019english, soland2021english}. 
Mirroring established classroom observation rubrics~\citep{pianta2008classroom} and drawing on empirically grounded observations, we develop a classification taxonomy for teacher attention that captures three key dimensions of instructional practice:  Content-focused instruction (discussing academic material), relationship-building (developing rapport and engagement), and session management (coordinating student behaviors and activities).

\section*{Findings}

\begin{figure*}[h]
\centering
\begin{subfigure}[b]{0.33\textwidth}
    \centering
    \includegraphics[width=\textwidth]{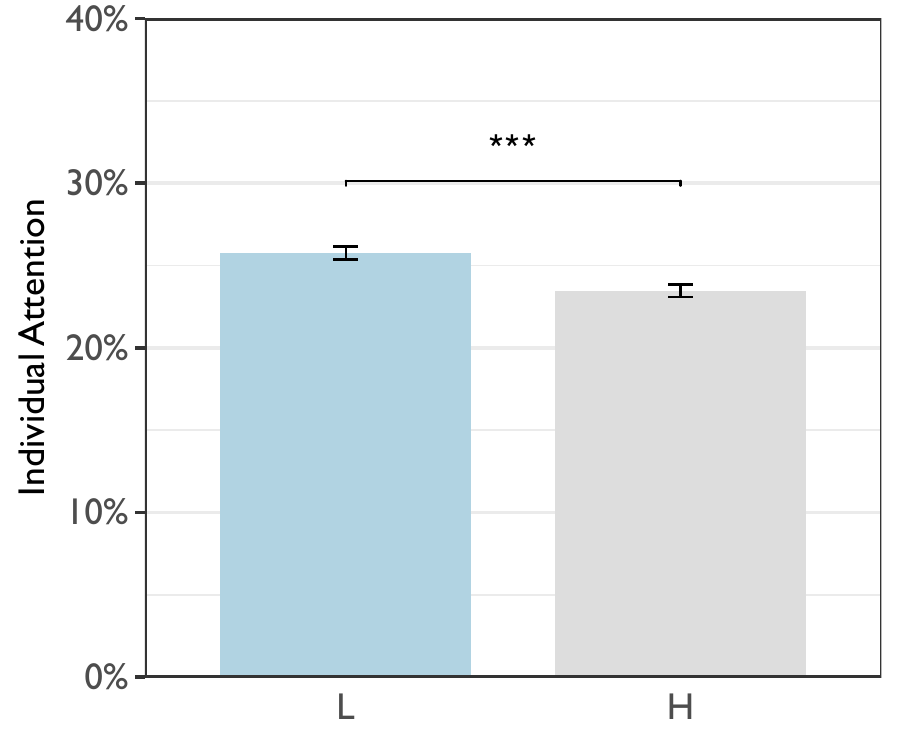}
    \caption{Main Plot}
    \label{fig:achievement_main}
    \end{subfigure}
    \hfill
    \begin{subfigure}[b]{0.66\textwidth}
    \centering
    \includegraphics[width=\textwidth]{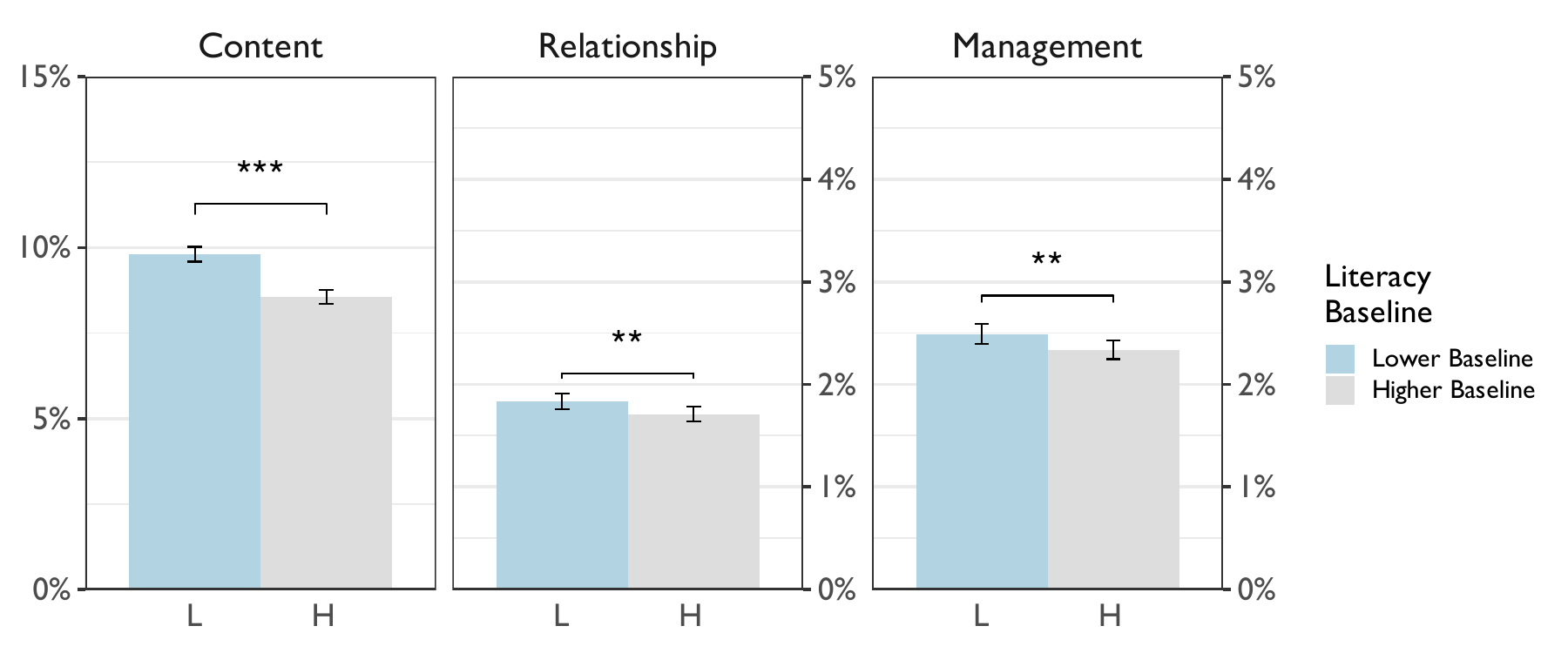}
    \caption{Topic Breakdown}
    \label{fig:achievement_topic}
    \end{subfigure}
\caption{\textbf{Individual attention allocation based on student achievement.} 
(a) Lower-achieving students (L) receive significantly more individual attention than their higher-achieving peers (H). 
(b) Breakdown of attention by type, showing that lower-achieving students receive more attention on content-related support, relationship-building interactions, and session management. 
Asterisks indicate statistical significance (*** $p < 0.001$, ** $p < 0.01$).}
\label{fig:achievement}
\end{figure*}

\subsection*{Study 1: Achievement-Based Patterns} 
Prior research provides evidence that educator attention can effectively bolster learning for lower-achieving students~\citep{lewis2012carino, soland2021english}. 
As a result, educators may consider students' achievement levels when determining how to allocate their attention across students and particularly focus on students who struggle with the material~\citep{brophy1970teachers, myhill2002bad}. 
We test whether educators provide more attention to lower-achieving students.
In our paired tutoring context, we define the lower-achieving student as the one with the lower baseline achievement score; see Appendix Section~\ref{app:data} on descriptive statistics of lower- and higher-achieving students. 
Figure~\ref{fig:achievement} reports the distribution of attention given to the lower-achieving and higher-achieving student within a tutoring pair, in terms of both (a) overall attention, and (b) the nature of attention. 

We find that educators are more attentive to students who have greater learning needs, with additional 2.3 percentage points (\ppm) attention to the lower-achieving student ($t(5248)=7.63$, $p\textless 0.001$).
This pattern holds across the types of attention with 1.25\pp for content, 0.12\pp for relationship and 0.16\pp for management (Figure~\ref{fig:achievement_topic}).
The most pronounced category is content-focused attention given to the lower-achieving student ($t(5248)=8.99$, $p\textless 0.001$). 
Overall, Study 1 suggests that educators are responsive to students' academic needs, providing them especially with more attention on instructional content.

\begin{figure*}[h]
    \centering
    \begin{subfigure}[b]{0.3\textwidth}
        \centering
        \includegraphics[width=\textwidth] {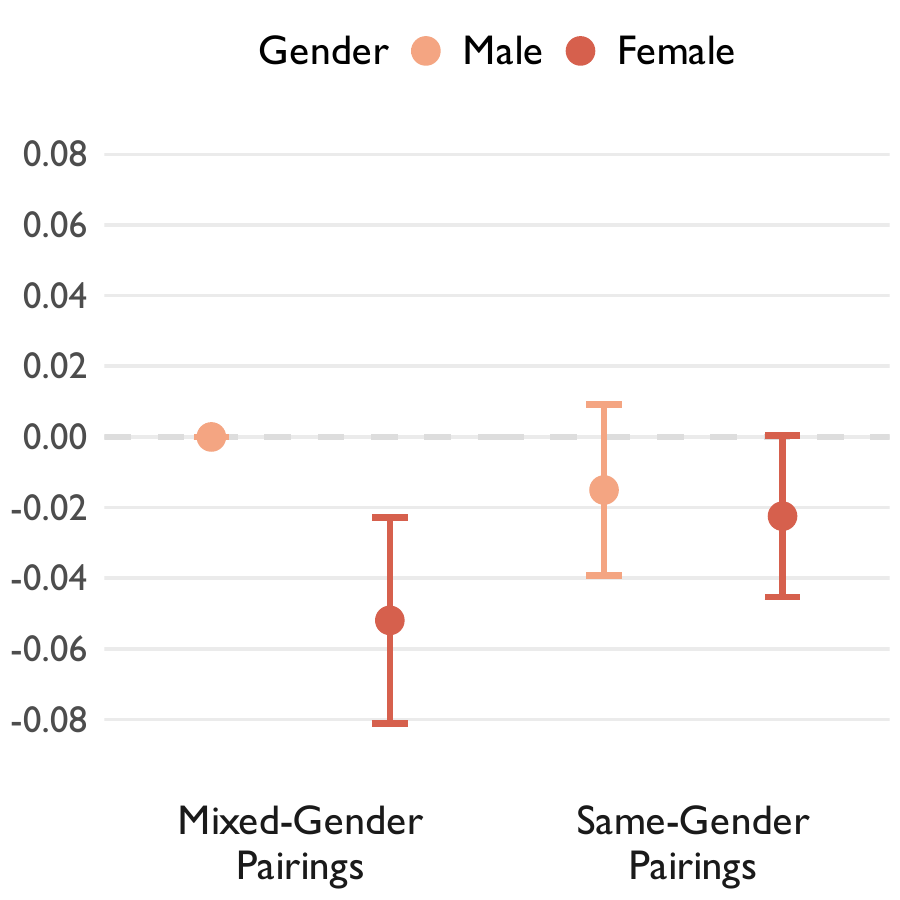}
        \caption{Gender Pairing}
        \label{fig:coef_gender}
    \end{subfigure}
    \begin{subfigure}[b]{0.3\textwidth}
        \centering
        \includegraphics[width=\textwidth] {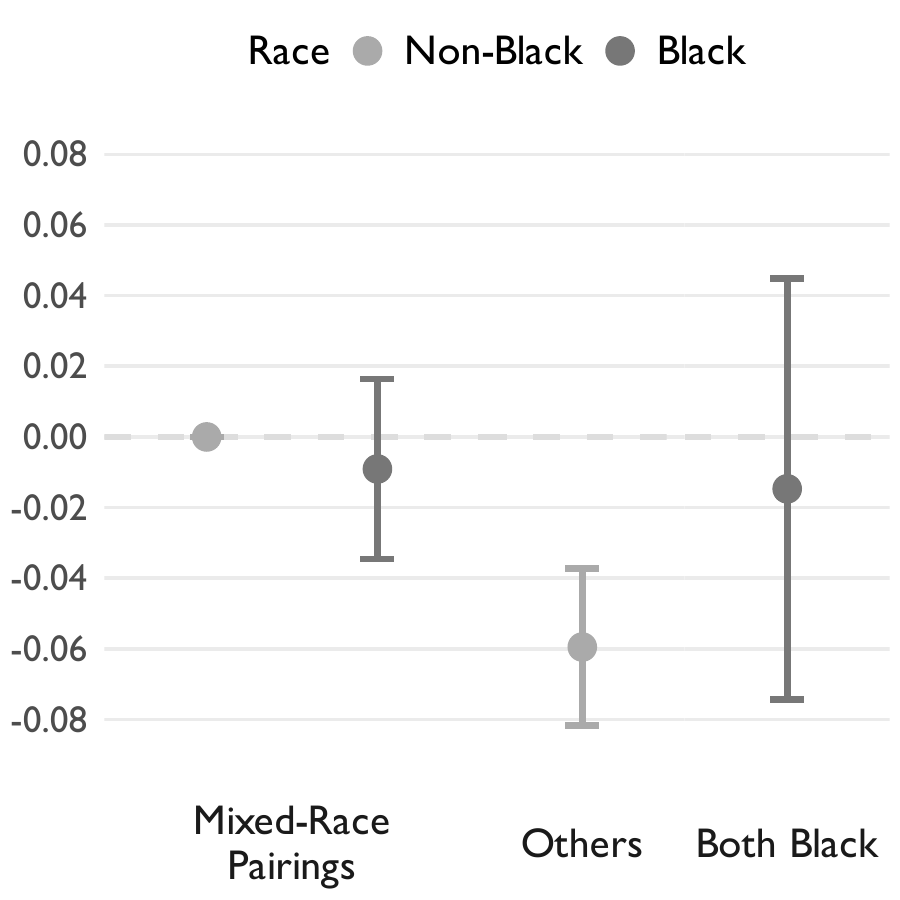}
        \caption{Race Pairing}
        \label{fig:coef_race}
    \end{subfigure}
    \begin{subfigure}[b]{0.3\textwidth}  
        \centering 
        \includegraphics[width=\textwidth]{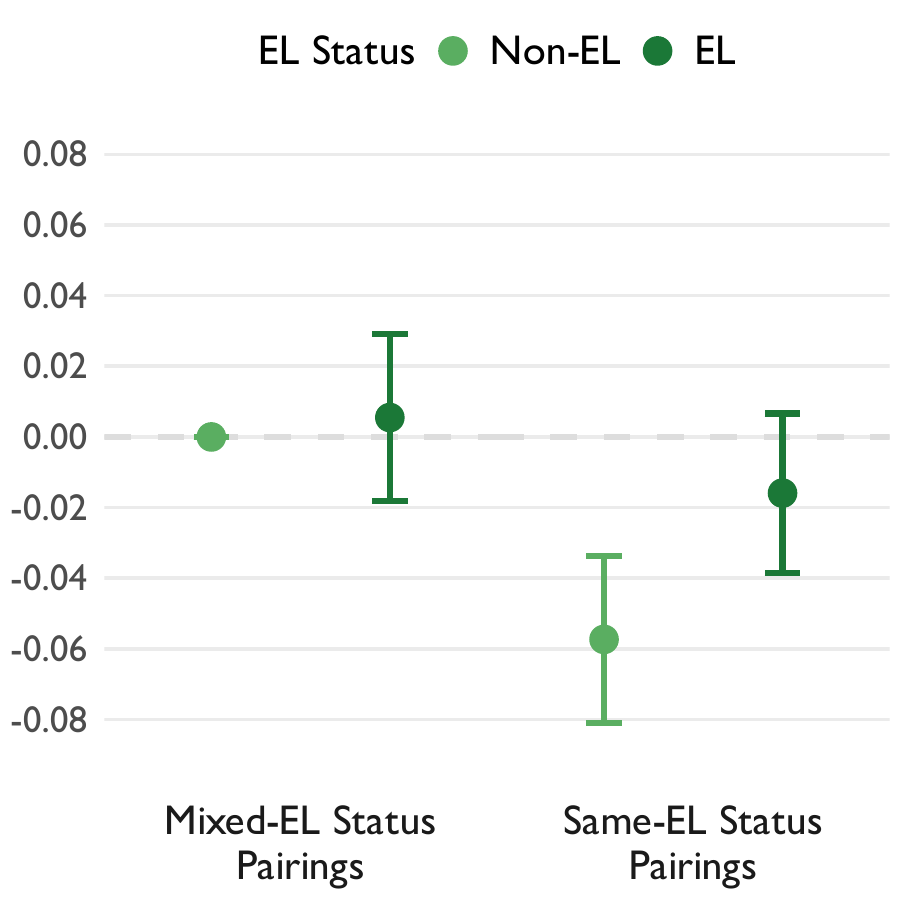}
        \caption{EL Status Pairing}
        \label{fig:coef_el}
    \end{subfigure}
    \caption{\textbf{Regression coefficients for individual attention allocation across student pairings.} The left-most dot is the reference group. Error bars indicate standard errors.
    (a) Gender pairings: In mixed-gender groups, female students receive significantly less attention than their male peers (the reference group). 
    (b) Race pairings: In groups with different races with a focus on Black students, Black students receive slightly more attention than their non-Black peer (the reference group).
    (c) EL status pairings: In groups with students of different EL status, we find that students get about the same amount of attention.}
    \label{fig:coef}
\end{figure*} 

\subsection*{Study 2: Demographic-Based Patterns} 
In addition to academic achievement, students' demographics---such as gender, race and home language---may also affect the attention they receive from educators. 
For example, ~\citet{gilliam2016early} conducted an eye-tracking study on preschool teachers and found that they tended to focus more on Black boys when asked to look for potential behavioral issues, even when no misbehavior occurred. 
Similarly, the demographic composition of student pairs---such as whether students are in same-gender or mixed-gender pairs---may further influence how educators distribute attention~\citep{kulik1982effects}.
While some differences in attention may stem from educators prioritizing lower-achieving students, others may persist beyond achievement and potentially reinforce structural disparities in educational opportunities.

To examine this in Study 2, we analyze how tutors allocate attention based on individual student demographics and the demographic composition of student pairs, while controlling for academic achievement. 
We focus on three key demographic indicators studied in prior work on educator attention: gender, race and EL status, as defined by the district and available in district administrative data. 
For gender, we use available classifications of male or female.
For EL status, we compare EL and non-EL students.
For race/ethnicity, we categorize students as Black or non-Black; 
while we have a more detailed measure of race and ethnicity, the sample size for non-Black, non-Hispanic students is too small to reliably estimate differences between each subgroup (see Appendix Section~\ref{app:data} for a sample breakdown). 
We focus on Black students because of the especially strong systemic barriers they often face, including economic disadvantage, lower test scores and low graduation rates~\citep{chetty2020race, national2024condition}.

Concretely, we model the amount of attention, $y_{ijk}$, a student $i$ receives when partnered with student $j$ in session $k$ as a function of: an indicator of their pair composition $I_{ijk}$ (\eg female in same-gender or mixed-gender pairings), their own baseline achievement $D_{i}$, their partner's achievement $D_{j}$ and  their relative achievement (lower or higher) $R_{ij}$. 
We cluster standard errors at the pair level, $e_{ij}$.

\begin{equation*}
   y_{ijk} = \beta_{1}I_{ijk} + \beta_{2}D_i + \beta_{3}D_{j} + \beta_{4} R_{ij} + e_{ij} 
\end{equation*}

Figure~\ref{fig:coef} reports $\beta_{1}$, the estimated differences in educator attention based on the student-level (gender, race, and EL status) and group demographics (same or mixed gender, race, and EL). 
We also report differences by the nature of attention below and include the plots in Appendix Section~\ref{app:add_results}.

Figure~\ref{fig:coef_gender} reports the amount of attention students of different genders and gender pairings receive.
Our findings highlight a pronounced gender gap in attention allocation. 
Female students receive significantly less individual attention than their male counterparts, with the largest disparity emerging in mixed-gender pairings, where female students receive 5.2 \pp less attention than their male partners [$\beta=-0.052(-0.081, -0.023)$, $p\textless 0.001$], even when controlling for achievement levels. 
Female students particularly receive less attention on content and management (rf. Appendix Section~\ref{app:add_results}).

Figure~\ref{fig:coef_race} reports the results for race, and  Figure~\ref{fig:coef_el} the results for EL status.
We find that when both students are non-Black or both are non-EL students, they receive significantly less individual attention compared to pairs where at least one student is Black or is an English learner.
Specifically, pairs of non-Black students receive 5.9 \pp less attention [$\beta=-0.059(-0.082, -0.037)$, $p\textless 0.001$], and pairs of non-EL students receive 5.7 \pp less attention [$\beta=-0.057(-0.081, -0.034)$, $p\textless 0.001$] compared to non-Black students and non-EL students in mixed pairings, respectively. 
This reduction is primarily driven by a overall decrease in content-focused attention (rf. Appendix Section~\ref{app:add_results}). 
Study 2 thus suggests that educators allocate disparate attention to students based on their demographics, beyond what is explained by students' measured achievement.

\subsection*{Study 3: Demographic Patterns by Achievement Relative to Peers} 
Studies 1 and 2 established two key findings: Educators generally prioritize the lower-achieving student in the pair (Study 1), but they also show systematic differences in attention allocation based on student demographics (Study 2). 
These findings raise a critical question: Does the tendency of educators to support the lower-achieving student of the group operate uniformly across demographics?

\begin{figure*}[h]
    \centering
    \begin{subfigure}[b]{0.58\columnwidth}
        \centering
        \includegraphics[width=\textwidth]{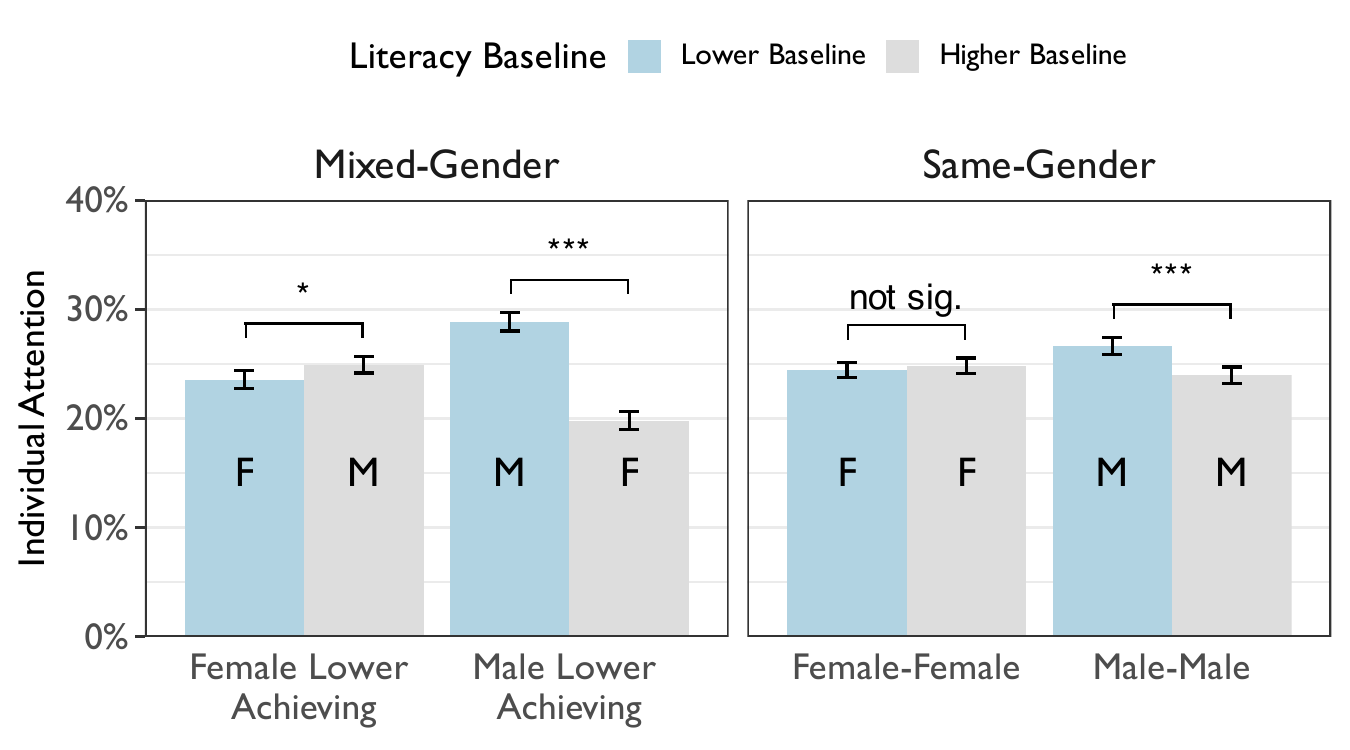}
        \caption{By Gender}
        \label{fig:interaction_gender}
    \end{subfigure}

    \begin{subfigure}[b]{0.49\columnwidth}
        \centering
        \includegraphics[width=\textwidth]{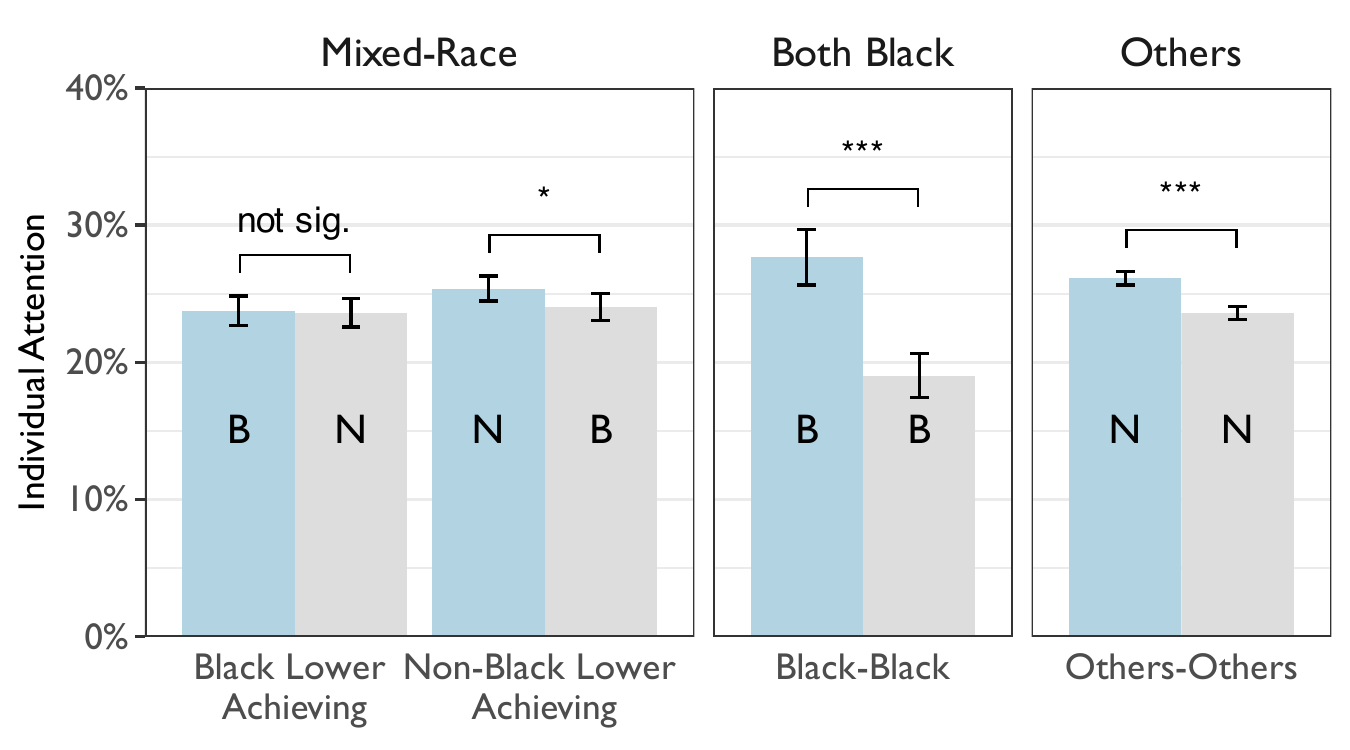}
        \caption{By Race (Black)}
        \label{fig:interaction_black}
    \end{subfigure}
    \begin{subfigure}[b]{0.49\columnwidth}
        \centering
        \includegraphics[width=\textwidth]{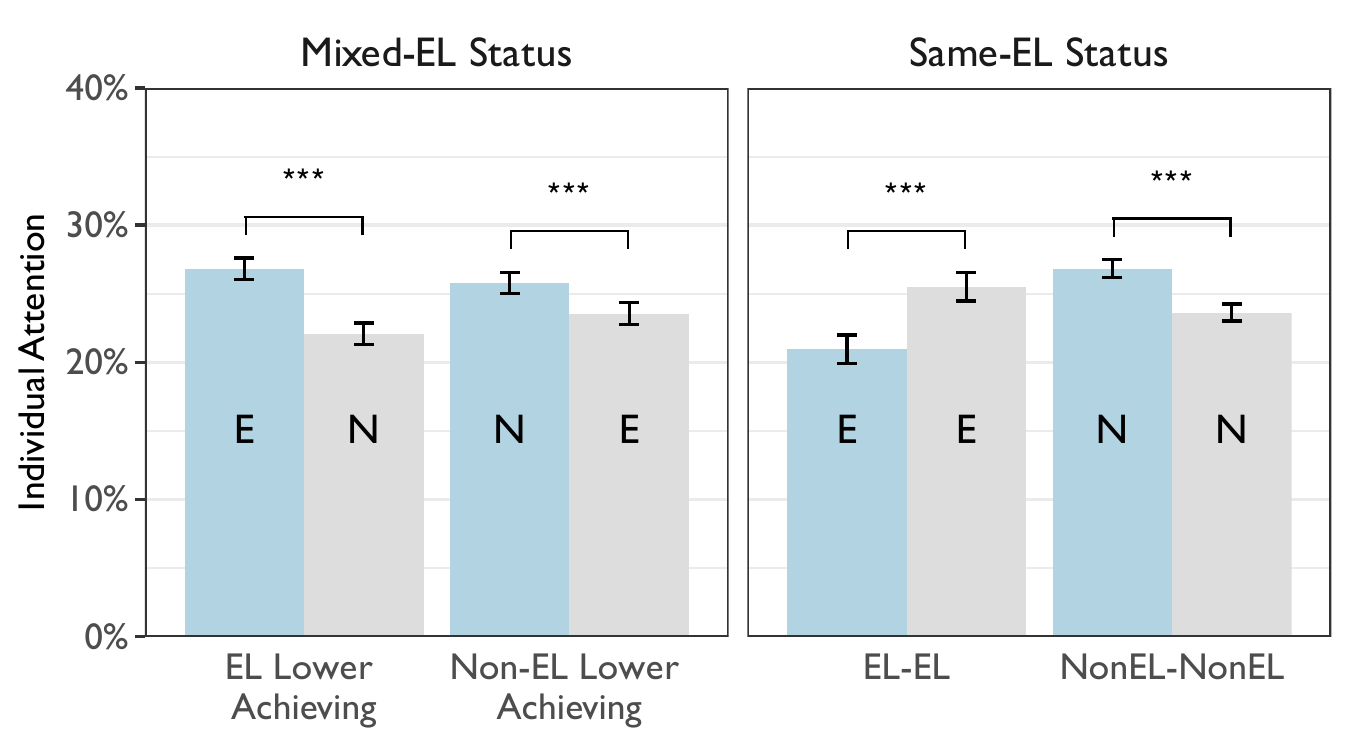}
        \caption{By English Learner Status}
        \label{fig:interaction_el}
    \end{subfigure}
    \caption{\textbf{Distribution of individual attention by achievement across student pairings.} Error bars indicate standard errors. 
    (a) Gender pairings: In mixed-gender groups, lower-achieving male students receive significantly more attention than their female counterparts. 
    (b) Race pairings: Lower-achieving Black students in Black-only groups receive significantly more attention, but this pattern does not hold in mixed-race pairings.
    (c) English Learner (EL) status pairings: Lower-achieving EL students receive less attention when paired with another EL student. 
    The lower-achieving non-EL student receives more attention in all settings. 
    }
    \label{fig:interaction}
\end{figure*}

To address this question, we examine whether a student's relative position as the lower- or higher-achieving student of their pair influences the attention they receive based on their demographics. 
For instance, when a Black student is the lower-achieving student of the pair, does the student receive the additional attention we would expect based on Study 1's findings? 
We report our results in Figure~\ref{fig:interaction} where we compare the amount of individual attention given to students when they are the lower- versus higher-achieving student of their pair, across different demographic groups.
We further analyze the differences by the nature of attention in Appendix Section~\ref{app:add_results}.

\paragraph{By Gender}
Figure~\ref{fig:interaction_gender} reveals a striking deviation from educators' general tendency to prioritize lower-achieving students. 
In mixed-gender pairs, lower-achieving female students not only fail to receive additional attention, but they receive 1.4 \pp \textit{less} attention than their higher-achieving male peers ($t(1279)=-2.32$, $p = 0.02$). 
This pattern stands in stark contrast to how educators respond to lower-achieving male students, who receive a substantial 9.1 \pp increase in attention compared to their higher-achieving female peers ($t(1262)=14.54$, $p \textless 0.001$). This increase is driven by all categories of attention including content-focused attention (rf. Appendix Section~\ref{app:add_results}). 

\paragraph{By Race}
Figure~\ref{fig:interaction_black} reveals another notable deviation from educators' typical pattern of prioritizing lower-achieving students. 
In mixed-race pairs, lower-achieving Black students do not receive the additional attention we would expect; they receive about the same amount of attention as their non-Black higher-achieving peer. 
This is a striking difference to when both students are Black, where the lower-achieving Black student receives 8.6 \pp more attention than their higher-achieving Black peer ($t(256)=7$, $p \textless 0.001$).
This increase in attention is also driven by all categories of attention (rf. Appendix Section~\ref{app:add_results}).

\paragraph{By EL Status}

Figure~\ref{fig:interaction_el} shows that in mixed EL pairs, the lower-achieving student receives more attention, following the expected pattern.
However, in pairs where both students have EL status, this trend reverses: the lower-achieving EL student receives 4.6 \pp \textit{less} attention than their higher-achieving EL peers ($t(740)=-5.56$, $p < 0.001$).
This stands in contrast to all other pairings, where lower-achieving students receive more attention.
The higher-achieving EL peer receives primarily more content-focused and relationship-building attention (rf. Appendix Section~\ref{app:add_results}).

\section*{Discussion}
Educator attention is one of the most powerful resources for improving student learning and well-being. 
Yet, educator attention is often subtle and difficult to study. 
In the past, researchers needed to manually observe and record observations to identify patterns with potentially meaningful consequences for students. 
The time costs of these observations often resulted in small samples, covering few educators and short time spans. 
The expansion of both large-scale virtual tutoring and computational methods for analyzing this data provides an unprecedented opportunity to understand patterns in educator-student interactions. 
Our study leverages these data and methods to complement existing observational insights, and provide new nuanced perspectives on educator attention.

Our findings reinforce some prior qualitative observations, such as gender-based disparities in educator attention~\citep{beaman2006differential}, and examine racial and EL disparities, areas that are understudied in large-scale settings.
Our work shows that educators tend to allocate more attention to lower-achieving students, consistent with the benefits of need-based attention.
However, we also identify notable deviations from this pattern.
Lower-achieving female students in mixed-gender pairs receive significantly less attention than their higher-achieving male peers, while lower-achieving male students receive substantially more attention than their higher-achieving female peers.
A similar deviation emerges for lower-achieving Black students who do not receive additional attention when paired with a non-Black student, but do receive it when both students are Black.
By contrast, higher-achieving EL students receive much more attention than their lower-achieving EL peers; one potential explanation for this pattern is that the tutor may find it easier to engage with the EL student who has the higher literacy level in these groups.
By examining these disparities in a large-scale setting, our study extends prior research and sheds light on the interaction of multiple factors, including student needs and demographics, that shape educator-student interactions.

Several limitations must be acknowledged. 
Our study focuses on educator utterances, excluding student talk, which limits our ability to examine how student-initiated behaviors shape educator responses.
Because educator-student interactions are inherently bidirectional~\citep{brophy1970teachers}, future work would benefit from incorporating student talk.
For example, the gender-based differences in educator attention from Study 2 and 3 may be partially explained by differences in student behaviors, such as boys exhibiting more attention-seeking behaviors~\citep{brophy1970teachers, irvine1986teacher, dobbs2004attention}.
Additionally, the absence of educator demographic information prevents us from studying how educator identities influence patterns in interaction---an important factor in instructional effectiveness~\citep{loeb2014good}. 

Despite these limitations, our study demonstrates the power of large-scale data and computational methods to uncover subtle but consequential patterns in educational interactions. 
By identifying disparities in educator attention, this work surfaces empirical evidence that informs efforts to improve teaching practices and create more equitable learning environments.

\section*{Materials and Methods}

\paragraph{Data and Processing.}
Our data is sourced from a U.S.-based early literacy, virtual tutoring provider that offers end-to-end services for school districts, including the tutoring platform, instructional materials, and tutors.
While the tutoring provider offers different forms of tutoring, our study uses the 2-on-1 tutoring sessions where a tutor conducts tutoring with two students. 
These students are in K-2 grades and were identified as being eligible for tutoring because they were performing below the early literacy benchmark.
The tutors are U.S. based, and are former classroom teachers and part-time teachers among others.
The tutoring interactions are video-based, integrated on the providers' online platform. 

Our study uses the tutor's audio recordings from the tutoring session, which were already individually saved from the students' audio recordings.
We excluded tutoring sessions where the students could not be matched to their metadata information, such as their gender or race. 
Tutors could opt-in to providing their background information including demographics, however not all did.
Therefore, we do not use the tutors' information in our analysis.
We transcribed the tutor's audio with an automatic speech recognition model ~\citep{bain2022whisperx}, where each transcribed utterance came with associated with start and end timestamps.
We manually checked the quality of the transcripts to validate their useability. 

The transcripts were de-identified by the tutoring provider, with their roster of student names. 
Because students are uniquely identifiable, their names are replaced with ``[Student A]'' and ``[Student B]'' in the transcripts.
We automatically trimmed transcripts to when both students are present in the session based on the entry timestamps of the students because our study is interesting in how the tutor attends to different students in the session.
Finally, we excluded sessions with durations that were less than half the planned tutoring session length because these sessions did not complete the desired amount of instruction. 
The final dataset size is \numTranscripts transcripts. 

\paragraph{Development of Attention Framework}
Our attention framework was developed through an empirically grounded approach, drawing on both prior research and manual observations. 
To determine the recipient of attention, we built on existing work in referee classification~\citep{ouchi2016addressee, jovanovic2004towards}, categorizing educator utterances based on the individual student.
Observing that many educator utterances engaged both students simultaneously, we explicitly included a ``both'' category. 
For the nature of attention, we leveraged prior work on educator attention~\citep[among others]{gunn2021measuring} and the training materials of the tutoring provider, both of which emphasize content instruction and relationship-building skills as central to interactions.
Additionally, upon manual observation, we noted a substantial number of educator utterances focused on session management, aligning with educational theory on instructional practice and classroom management literature~\citep[among others]{darling2007preparing}. 
These observations informed our framework, ensuring it captured key theoretical and empirical dimensions of educator-student interactions.

\paragraph{Human Annotation of Recipient of Attention.}
A subset of \numAnnotations utterances were sampled from the corpus for annotation. 
Utterances were sampled with the constraint that at least 10 prior utterances as preceding context.
We recruited 4 annotators, including two of the authors, who were trained on our attention framework and familiar with the data context to annotate for both the recipient and nature of attention. 
An additional category of ``NA'' if annotators were unsure about how to classify the recipient of the utterance.  
An initial round of 50 examples was annotated to assess inter-rater agreement, yielding Fleiss $\kappa = 0.66$ averaged over all categories.
While this indicated substantial agreement \citep{landis1977measurement}, we had some cases marked with ``NA'' due to missing contextual cues (\eg tone of voice) available in the original audio recordings.
Therefore, we decided to doubly annotate all \numAnnotations examples and resolve disagreement manually. 
Before adjudication, the final agreement score was Cohen's $\kappa=0.613$.
The two annotation authors resolved examples with disagreements and marked with ``NA'' by listening to the original audio recordings.  

\paragraph{Human Annotation of Nature of Attention.}
The same subset of \numAnnotations utterances was used to annotate the nature of attention. 
We employed a Human+LLM annotation approach, leveraging recent advances in combining human expertise with AI to scale annotation efforts~\citep[among others]{wang2024human}. 
We first prompted an LLM to annotate the dataset using our taxonomy, after which two co-authors reviewed and corrected the labels.
Human annotators played a particularly important role to ensure that the final labels captured the nature of educator attention in the utterances.

\paragraph{Computational Annotation of Utterances}
To automatically label tutor utterances with the recipient and nature categories at scale, we split our annotated dataset into a 7:1:2 train:validation:test set and fine-tuned a RoBERTa-large model on the training examples \citep{liu2019robertaD}. 
We created new token separators to separate the pretext context (10 preceding lines) and target utterance in the input sequences, formatting them as ``[PRETEXT\_TOKEN] \{pretext\} [TARGET] \{target\}'' to match what annotators saw during labeling.
We used a class-balanced loss function \citep{cui2019class} during fine-tuning, as some categories appeared more frequently than others. 
We performed a hyper-parameter search over the class-balanced loss annealing term $\beta$ and learning rate $\alpha$, fine-tuning for 20 epochs under each configuration.
The best model was selected based on its validation F1 score which was trained with with $\beta=0.9$ and $\alpha=5e6$. 
This final SoS classification model achieved a test F1 score of $71.8$, and test accuracy of $72.5$. 
We applied this model to automatically label all tutor utterances in our larger corpus of \numUtterances utterances for the downstream analyses in this work.

We compare the performance of our fine-tuned classifier to other modeling alternatives, including heuristic-based methods (\eg detecting the presence of a student’s name) and k-shot prompting with large language models like Claude and GPT-4, and find that our classifier outperforms most methods. 
Additionally, our classifier offers a more cost-effective solution while maintaining high accuracy. 
For further details on our model evaluations, please see Appendix Section~\ref{app:add_model}.

\section*{Acknowledgments}
We are grateful to our partners in this research, On Your Mark and Uplift Education.
This research was graciously supported by the Overdeck Family Foundation, Smith Richardson Foundation, and Arnold Ventures.
We also thank Lena Phalen and Professor Jennifer Osuna for insightful conversations.

\bibliography{custom}

\appendix
\onecolumn

\section{Data \label{app:data}}

\subsection{Context}
Our data is sourced from a U.S.-based early literacy, virtual tutoring provider that offers end-to-end services for school districts, including the tutoring platform, instructional materials, and tutors.
While the tutoring provider offers different forms of tutoring, our study uses the 2-on-1 tutoring sessions where a tutor conducts tutoring with two students. 
Each tutoring session is scheduled to be 20 minutes and the tutoring provider prioritizes consistent tutor assignments for students.
The tutoring sessions take place from the latter half of the academic year.
The research team executed Data Use Agreements with the tutoring provider and a charter school network that outlined the allowable usage of the data to improve instruction in collaboration with an educational agency. 
Following FERPA guidelines, we were eligible to engage in secondary data analysis with student data, which is what we did for this study.
The tutoring curriculum includes a focus on phonics, phonological awareness, and fluency. 
These students are in K-2 grades and were identified as being eligible for tutoring because they were performing below the early literacy benchmark.
The tutors are U.S. based, and are former classroom teachers and part-time teachers among others;
they receive training and professional development, focused on content
knowledge, building relationships, and effective delivery of the intervention. 
The tutoring interactions are video-based, integrated on the providers' online platform. 
Additional demographic information on students and tutors can be found in Table~\ref{tab:demographics}.

\begin{table}[t]
  \centering
    \begin{tabular}{@{}lr@{}}
    \toprule
        \multicolumn{2}{c}{\textbf{Tutors}} \\ 
    \midrule
        \# of Tutors           & 125    \\
        Demographics           &  \\
        \quad Asian American   & 6\%    \\
        \quad Black            & 35\%   \\
        \quad Latina/o/x       & 12\%   \\
        \quad Multiracial      & 8\%   \\
        \quad White            & 45\%  \\
        Educational Status \\
        \quad Grad Studying    & 8\%         \\
        \quad College Grad     & 47\%\\
        \quad College Studying & 13\%   \\
        \quad High School Grad & 6\%   \\
        Teaching Experience & \\
        \quad Former Teacher   & 33\%        \\
        \quad Part-time Teacher  & 19\%  \\
        \quad Other Education Professional & 21\% \\ 
        \bottomrule
    \end{tabular} 
    \hspace{2em}
    \begin{tabular}{@{}lr@{}}
    \toprule
    \multicolumn{2}{c}{\textbf{Students}} \\ 
    \midrule
        \# of Students        & 757      \\
        Grade                 &   \\
        \quad Kindergarten    & 207 (27\%)  \\
        \quad First           & 293 (39\%)   \\
        \quad Second          & 257 (34\%)   \\
        Demographics          &  \\
        \quad Female          & 53\%        \\
        \quad Asian American  & 1\%        \\
        \quad Black           & 22\%         \\
        \quad Latina/o/x      & 70\%        \\
        \quad Multiracial     & 3\%        \\
        \quad White           & 4\%        \\
        \quad Multilingual Learners   & 35\%        \\
        \quad Student with Disabilities  & 5\%        \\
        \quad LIEM           & 89\%        \\ 
    \bottomrule
  \end{tabular}
  \vspace{1em}
  \caption{\textbf{Statistics on identifiable teachers and students.} Tutor demographics are reported for those who could be matched to transcripts using the platform ID crosswalk (n = 89 out of 125). 
  For each demographic, we use the naming convention from \cite{robinson2024effects}. LIEM = Low-income economically marginalized.
  \label{tab:demographics}
  }
\end{table}

\begin{table}[t]
\centering
    \begin{tabular}[t]{lrrr}
    \toprule
    \textbf{Measure} & \textbf{Count} & \textbf{Mean} & \textbf{Standard Deviation} \\
    \midrule
    Session Duration (in seconds) & $5,604,605$ & $1067.75$ & $180.68$ \\
    Number of Words & $6,723,347$ & $1280.88$ & $442.89$ \\
    Number of Utterances & $1,157,970$ & $220.61$ & $71.74$ \\
    \bottomrule
    \end{tabular}
    \vspace{1em}
    \caption{\textbf{Descriptive statistics of transcripts.}}
    \label{tab:transcript}
\end{table}

\subsection{Transcription and Metadata}
The tutor's audio from 2:1 sessions was transcribed with WhisperX \cite{bain2022whisperx}, associated with timestamps.
The students' audio suffered from low audio quality and was not transcribed: tutoring happened in school, therefore the loud background noise made transcription unreliable.
Each student is linked with metadata on their demographic information, which includes information about the student's gender, race, English language learner (ELL) status, socio-economic status, and special education status. 
Each student is also linked with their mid-of-year (MOY) and end-of-year (EOY) Dynamic Indicators of Basic Literacy Skills (DIBELS) performance, which is a measure of the student's  literacy skills.
The MOY DIBELS examination was taken prior to the tutoring sessions we use for this study; this acts as a control on the student's baseline performance, which may impact how much attention they receive from the tutor. 
We excluded tutoring sessions where the students could not be matched to their metadata information. 
Tutors could opt-in to providing their background information including demographics, however not all did.
Therefore, we do not use the tutors' information in our analysis. Table \ref{tab:transcript} reports descriptive statistics on sessions and transcripts. 

\subsection{Preprocessing} 
The transcripts were de-identified by the tutoring provider, with their roster of tutor and student names. 
Because students are uniquely identifiable, their names are replaced with ``[Student A]'' and ``[Student B]'' in the transcripts.
Because we're interested in how the tutor talks with more than one student in the session, we trimmed transcripts to when both students are present in the session; we automatically do this using the entry timestamps of the students.
Finally, we excluded sessions with durations that were less than half the planned session length (10 minutes).
The final dataset size is $\sim$ 5.2k transcripts. 

\subsection{Lower- vs. Higher-Achieving Students}
In our paired tutoring context, we define the lower-achieving student as the one with the lower baseline achievement score. 
This baseline is determined by the student's literacy test score, specifically the DIBELS composite score, taken prior to data collection \citep{good2001using}. 

Table \ref{tab:dibel_status} shows the statistics of the baseline achievements for the lower-achieving student and the higher-achieving student in standard deviation units. 
The DIBELS composite scores are standardized within each grade since the DIBELS composite score, which combines multiple subtests (e.g., phonemic awareness, oral reading fluency) varies by expected skill level per grade. 

\begin{table}[h]
\centering
    \begin{tabular}[t]{lrr}
    \toprule
    \textbf{Achievement Status} & \textbf{Mean} & \textbf{Standard Deviation} \\
    \midrule
    Higher & $0.48$ & $0.95$\\
    Lower & $-0.35$ & $0.81$\\
    \bottomrule
    \end{tabular}
    \vspace{1em}
    \caption{\textbf{Descriptive statistics of the lower- vs. higher-achieving student in each tutoring pair.}}
    \label{tab:dibel_status}
\end{table}

\section{Additional Attention Results \label{app:add_results}}

\subsection{Nature of Attention from Study 2}
The main study reports the total amount of attention students received by their demographic background and pair composition receive. 
Here, we report the amount of attention broken down to our three main categories of attention: Content, relationship-building, and management. 
We use the same model as in Study 2 of our paper, included here for reference: 
We model the amount of that attention type, $y_{ijk}$, a student $i$ receives when partnered with student $j$ in session $k$ as a function of: an indicator of their pair composition $I_{ijk}$ (\eg female in same-gender or mixed-gender pairings), their own baseline achievement $D_{i}$, their partner's achievement $D_{j}$ and their relative achievement (lower or higher) $R_{ij}$. 
We cluster standard errors at the pair level, $e_{ij}$. 

\begin{equation*}
   y_{ijk} = \beta_{1}I_{ijk} + \beta_{2}D_i + \beta_{3}D_{j} + \beta_{4} R_{ij} + e_{ij} 
\end{equation*}

\begin{figure*}[h]
    \centering
    \begin{subfigure}[b]{0.3\columnwidth}
        \centering
        \includegraphics[width=\textwidth]{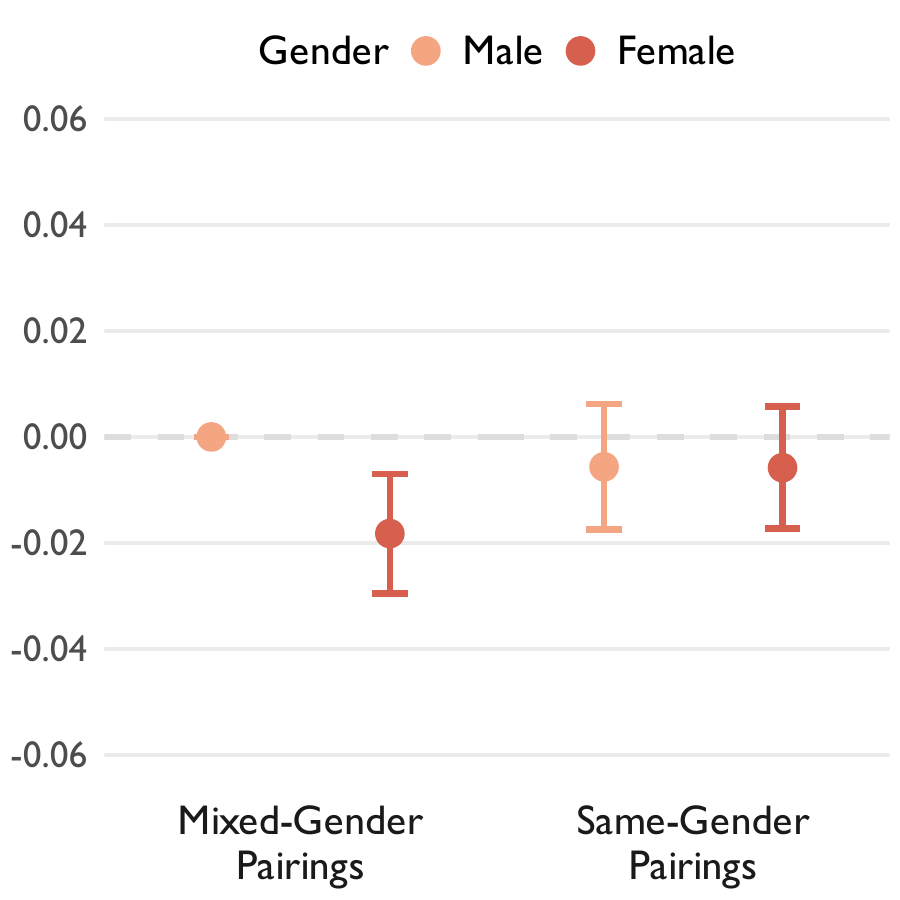}
        \caption{Content}
    \end{subfigure}
        \begin{subfigure}[b]{0.3\columnwidth}
        \centering
        \includegraphics[width=\textwidth]{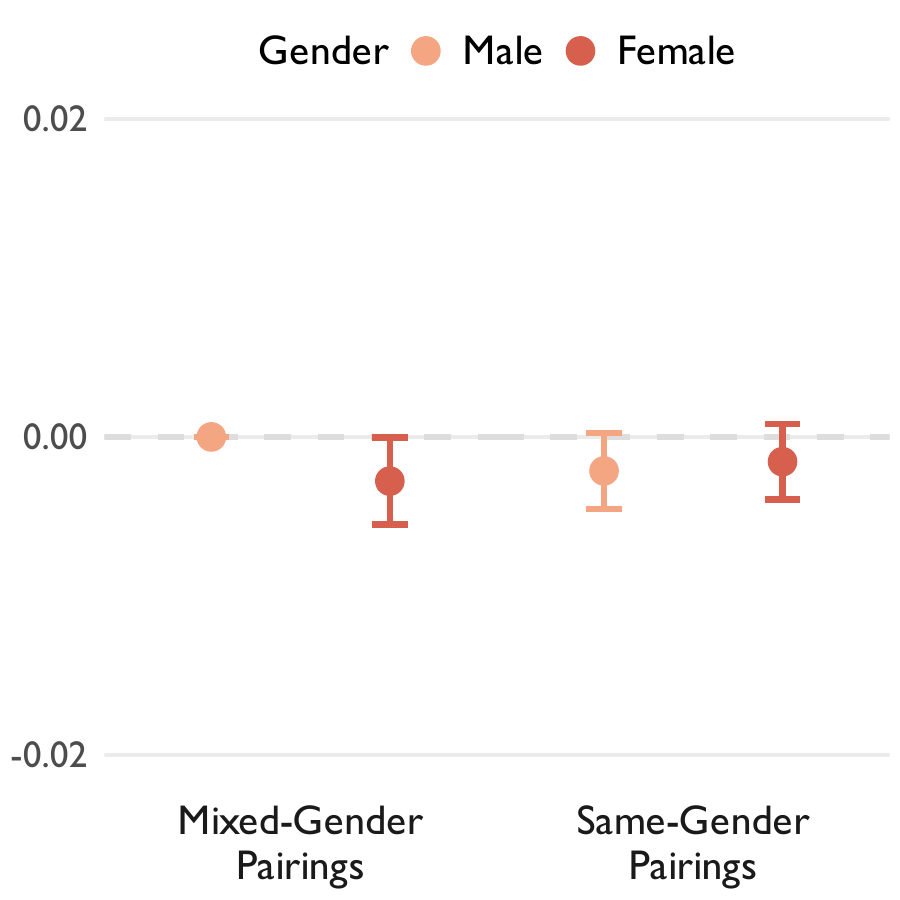}
        \caption{Relationship}
    \end{subfigure}
        \begin{subfigure}[b]{0.3
        \columnwidth}
        \centering
        \includegraphics[width=\textwidth]{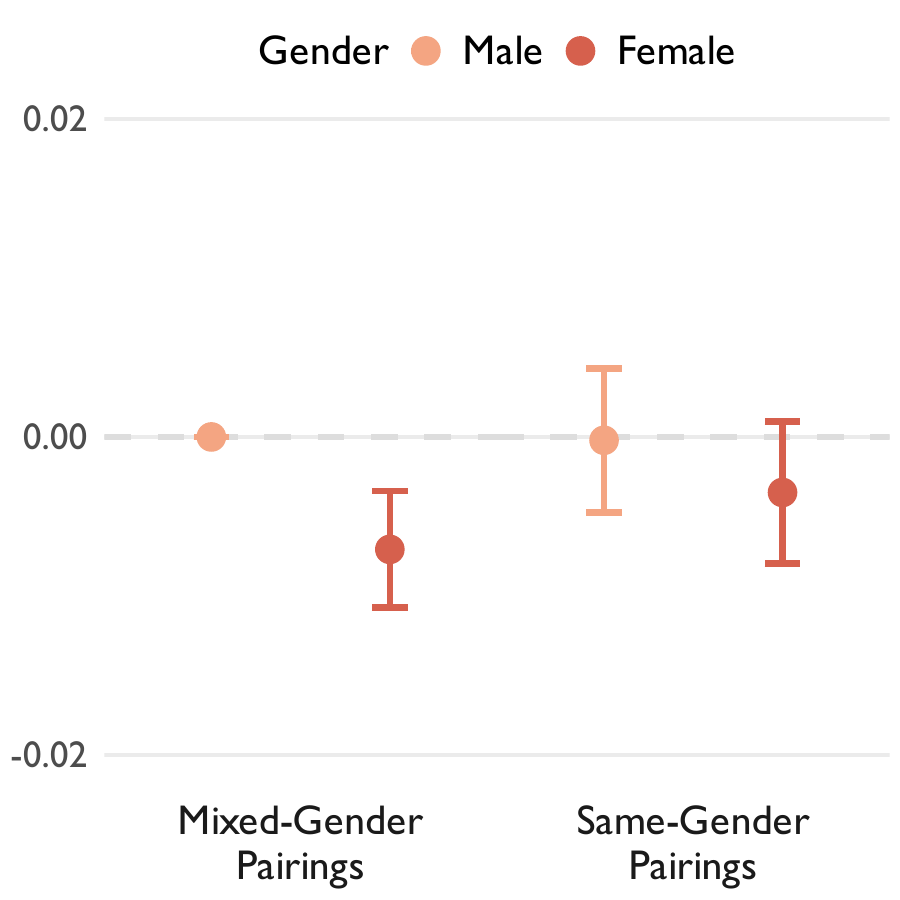}
        \caption{Management}
    \end{subfigure}
    \caption{\textbf{Nature of attention by gender.}
    }
    \label{fig:study2_gender}
\end{figure*}

Figure~\ref{fig:study2_gender} reports the estimated differences in educator attention based on gender. 
We find that female students in mixed-gender pairings receive much less content-focused and management-focused attention, than male students.

\begin{figure*}[h]
    \centering
    \begin{subfigure}[b]{0.3\columnwidth}
        \centering
        \includegraphics[width=\textwidth]{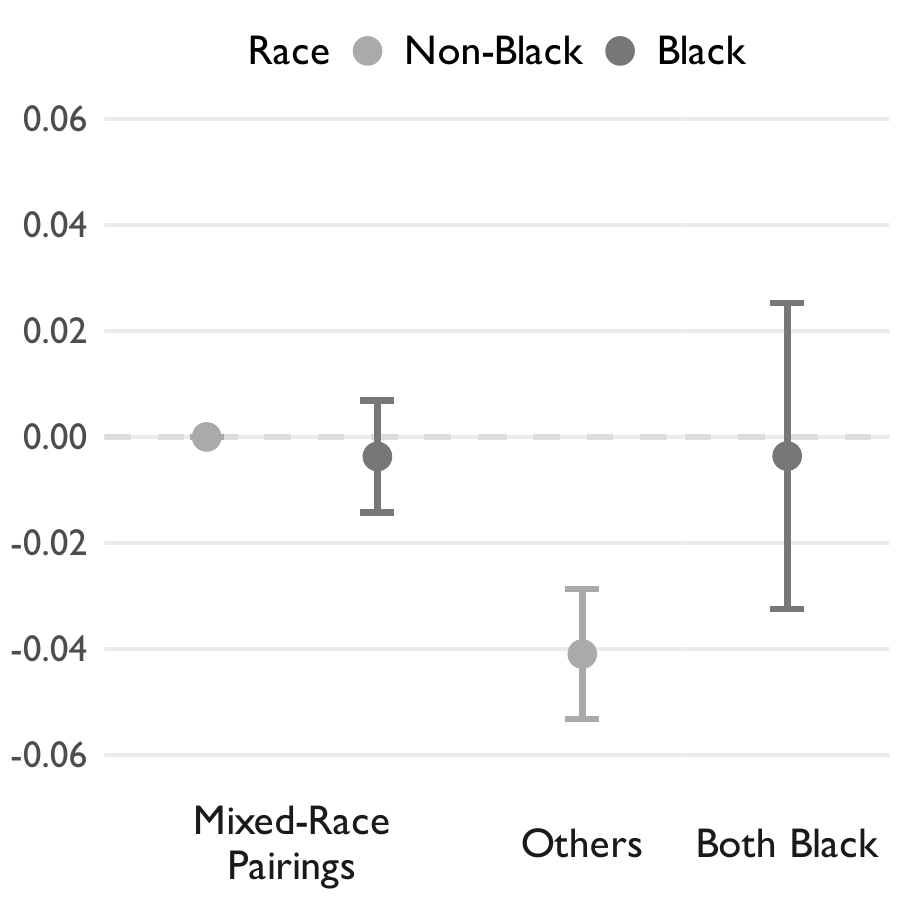}
        \caption{Content}
    \end{subfigure}
        \begin{subfigure}[b]{0.3\columnwidth}
        \centering
        \includegraphics[width=\textwidth]{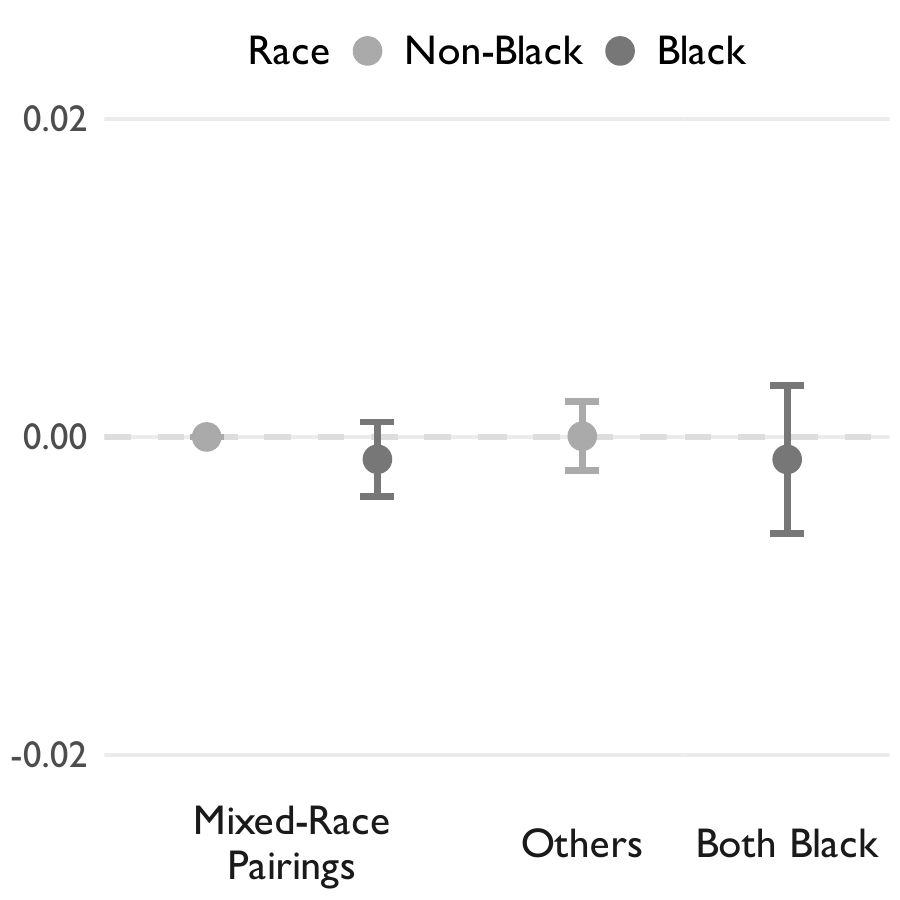}
        \caption{Relationship}
    \end{subfigure}
        \begin{subfigure}[b]{0.3
        \columnwidth}
        \centering
        \includegraphics[width=\textwidth]{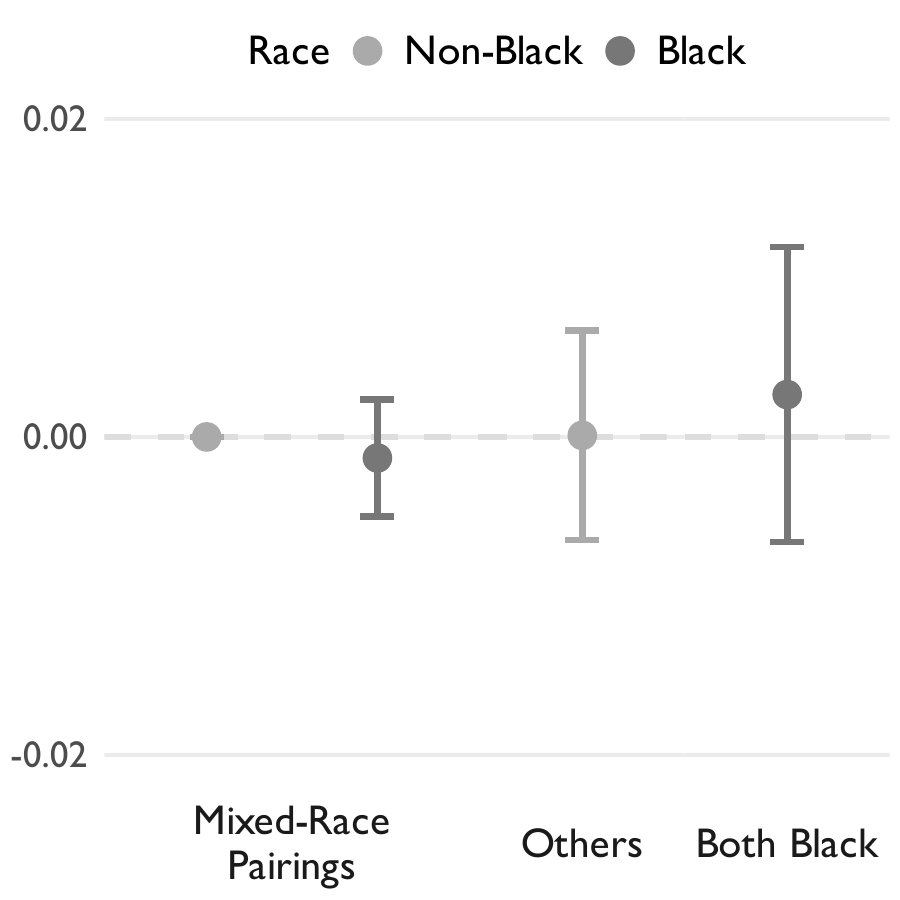}
        \caption{Management}
    \end{subfigure}
    \caption{\textbf{Nature of attention by race.}
    }
    \label{fig:study2_race}
\end{figure*}

Figure~\ref{fig:study2_race} reports based on race. 
We find that pairs of non-Black students receive much less individual attention on content. 

\begin{figure*}[h]
    \centering
    \begin{subfigure}[b]{0.3\columnwidth}
        \centering
        \includegraphics[width=\textwidth]{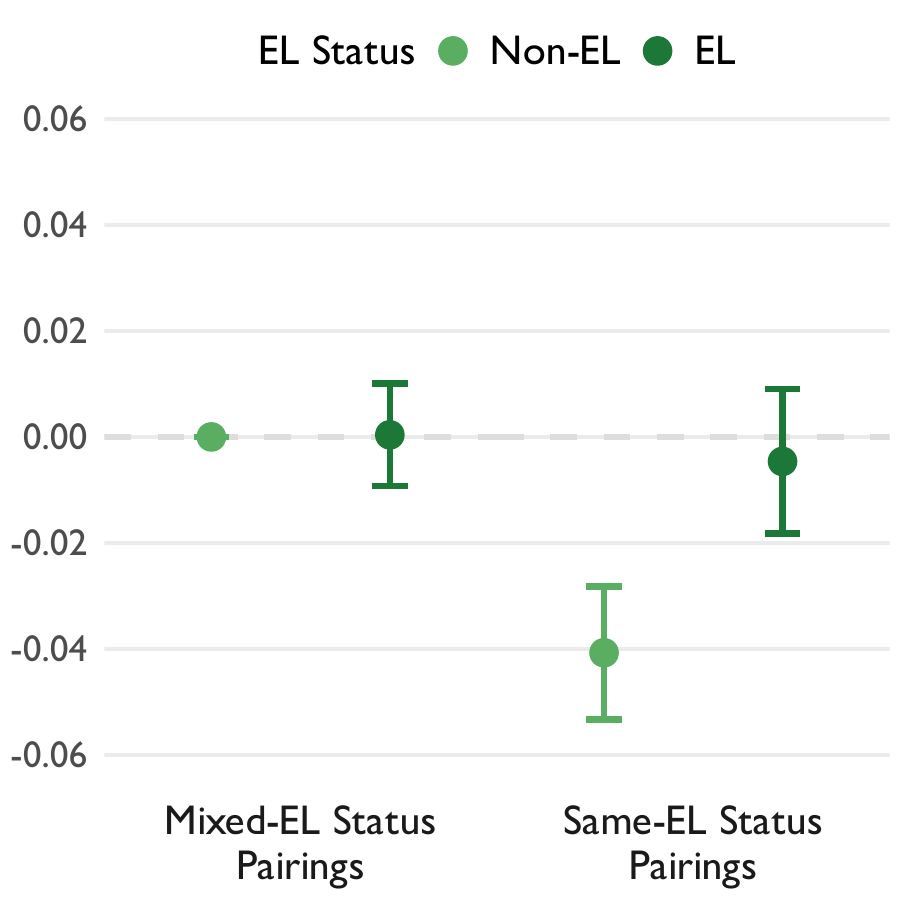}
        \caption{Content}
    \end{subfigure}
        \begin{subfigure}[b]{0.3\columnwidth}
        \centering
        \includegraphics[width=\textwidth]{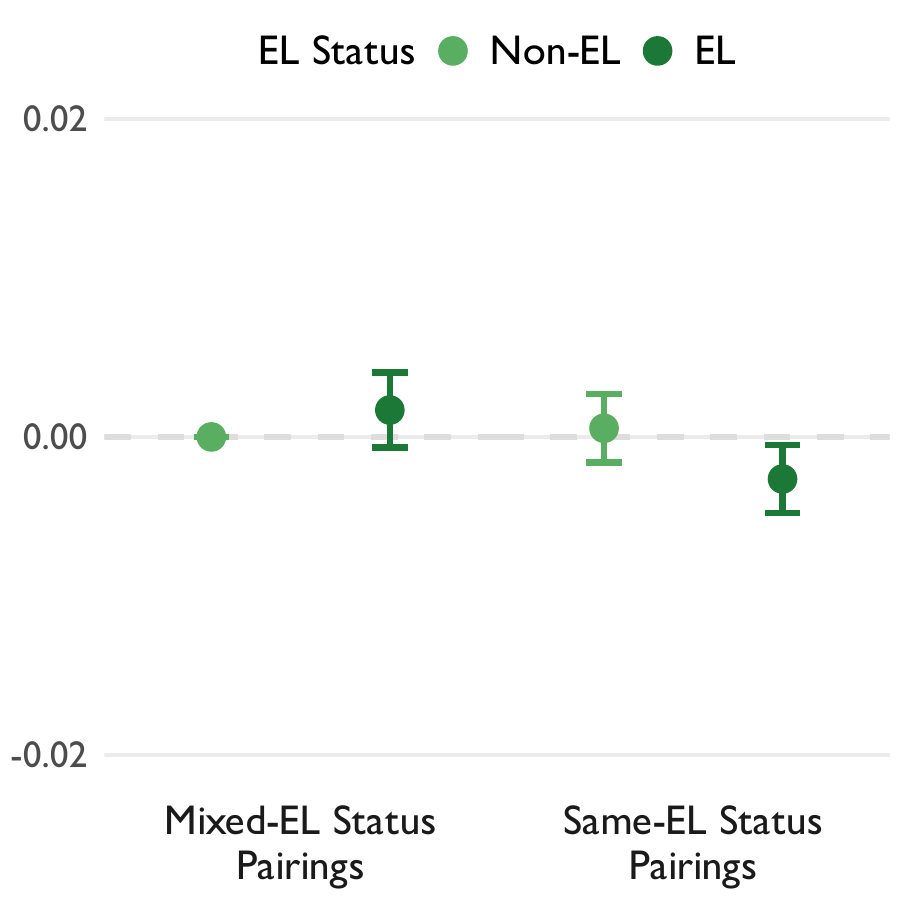}
        \caption{Relationship}
    \end{subfigure}
        \begin{subfigure}[b]{0.3
        \columnwidth}
        \centering
        \includegraphics[width=\textwidth]{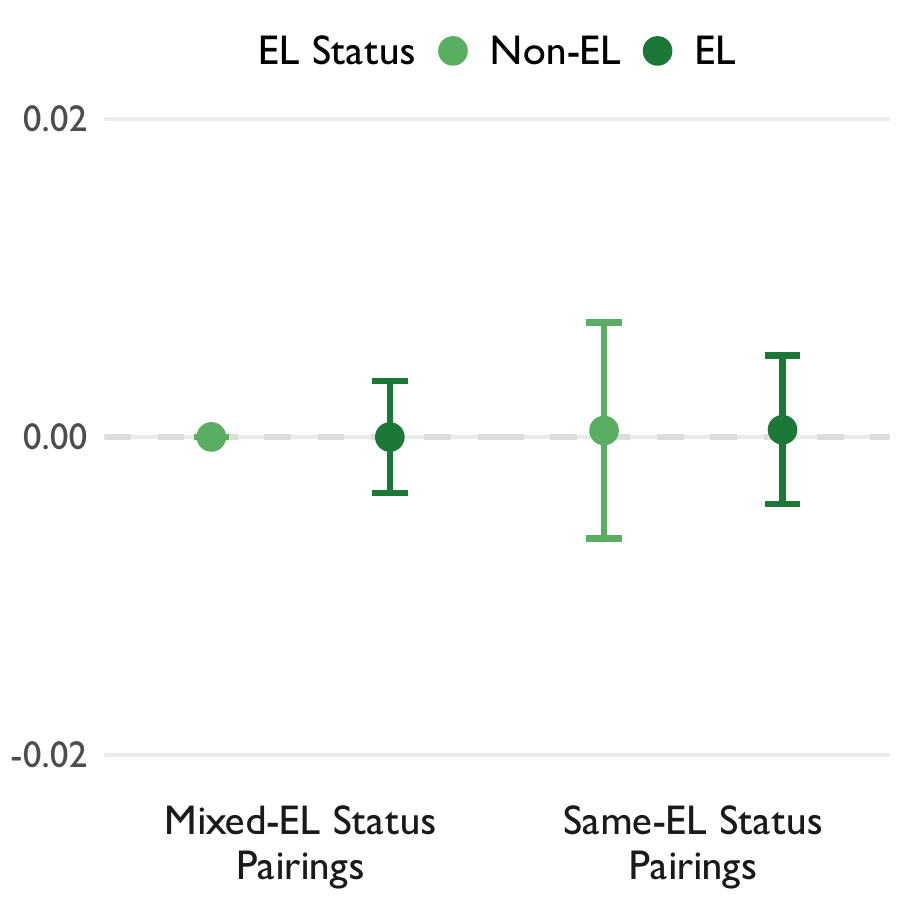}
        \caption{Management}
    \end{subfigure}
    \caption{\textbf{Nature of attention by EL status.}
    }
    \label{fig:study2_el}
\end{figure*}

Figure~\ref{fig:study2_el} reports based on the students' EL status. 
We find that pairs of non-EL students receive much less individual attention on content as well.  

\newpage 
\subsection{Nature of Attention from Study 3}
\begin{figure*}[h]
    \centering
    \begin{subfigure}[b]{0.3\columnwidth}
        \centering
        \includegraphics[width=\textwidth]{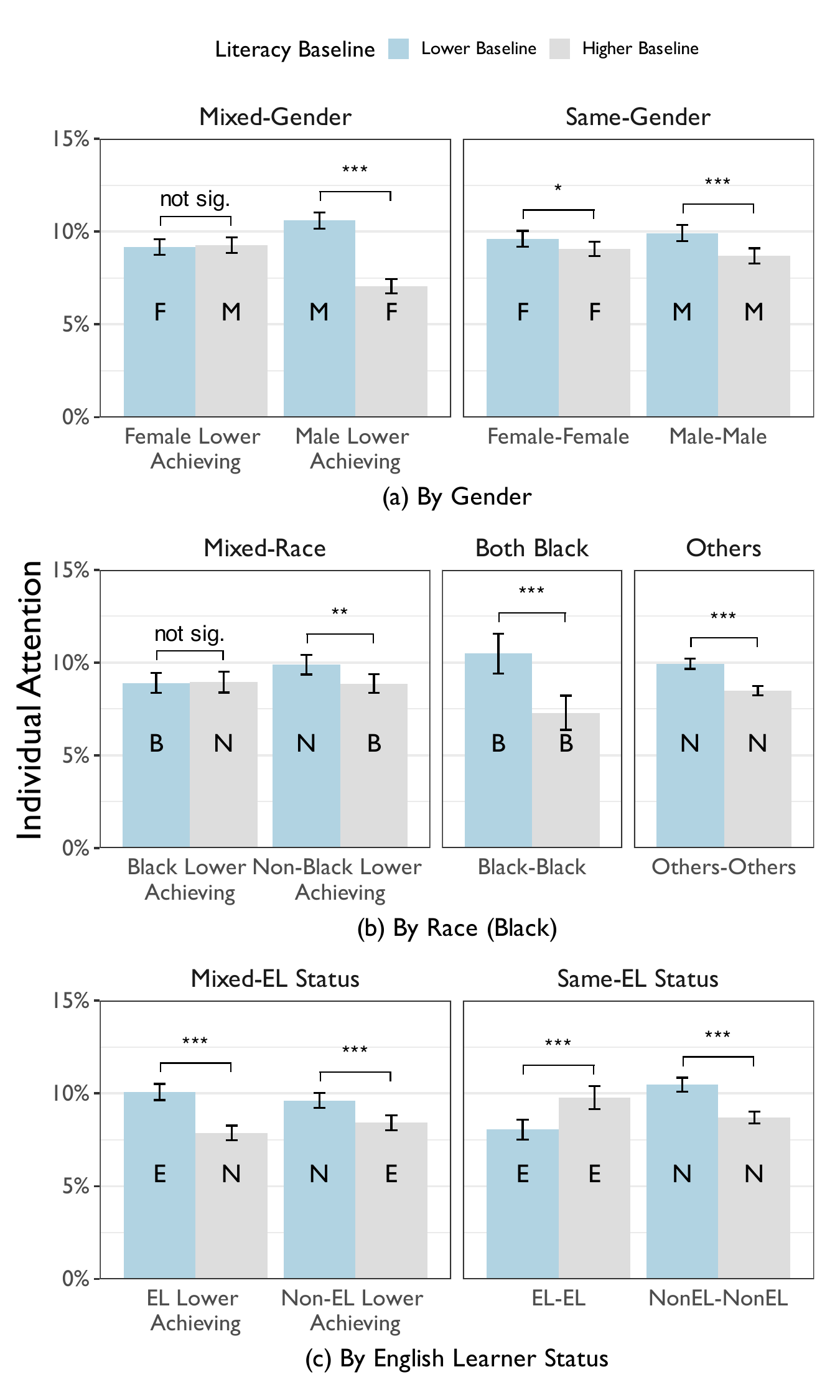}
        \caption{Content}
        \label{fig:study3_content}
    \end{subfigure}
        \begin{subfigure}[b]{0.3\columnwidth}
        \centering
        \includegraphics[width=\textwidth]{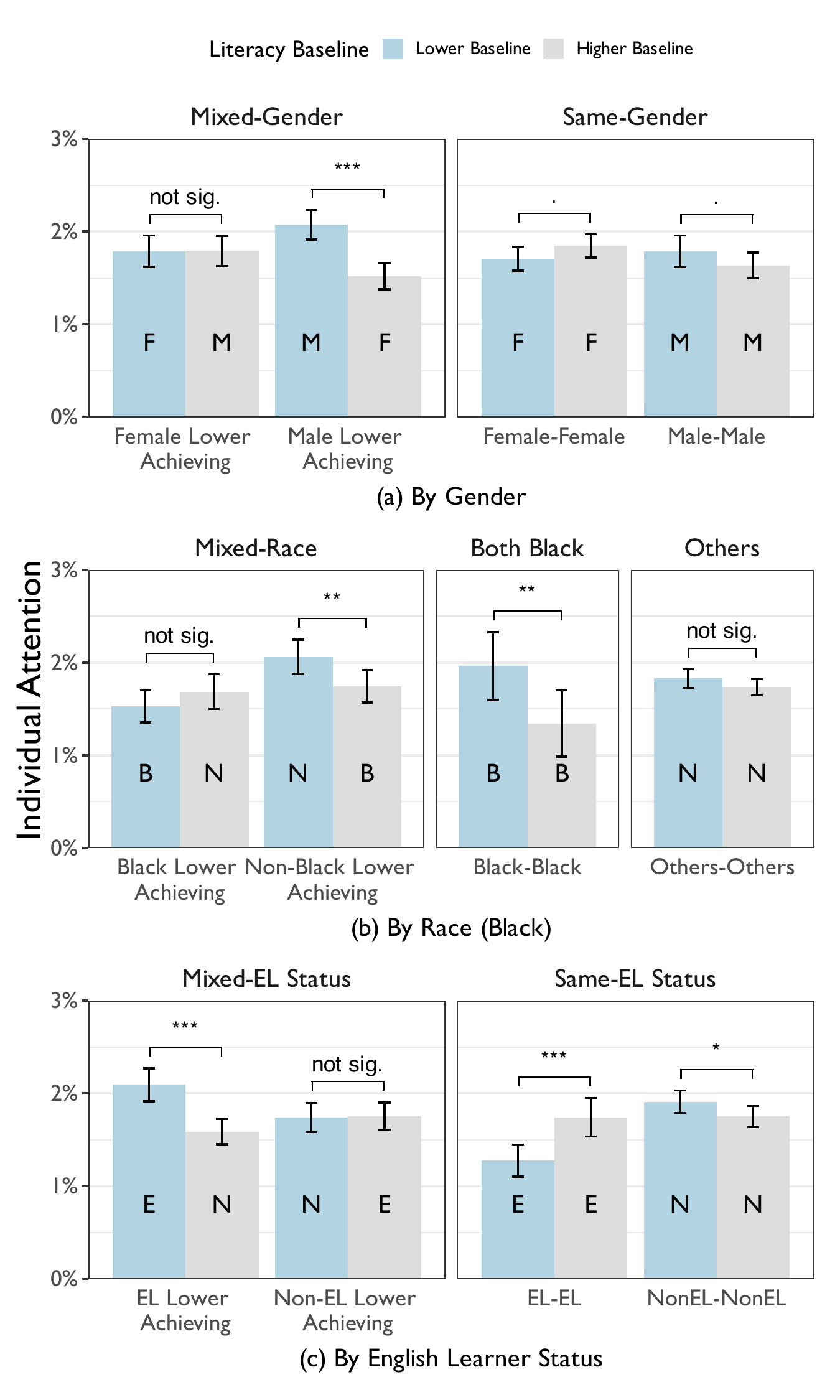}
        \caption{Relationship}
        \label{fig:study3_rel}
    \end{subfigure}
        \begin{subfigure}[b]{0.3
        \columnwidth}
        \centering
        \includegraphics[width=\textwidth]{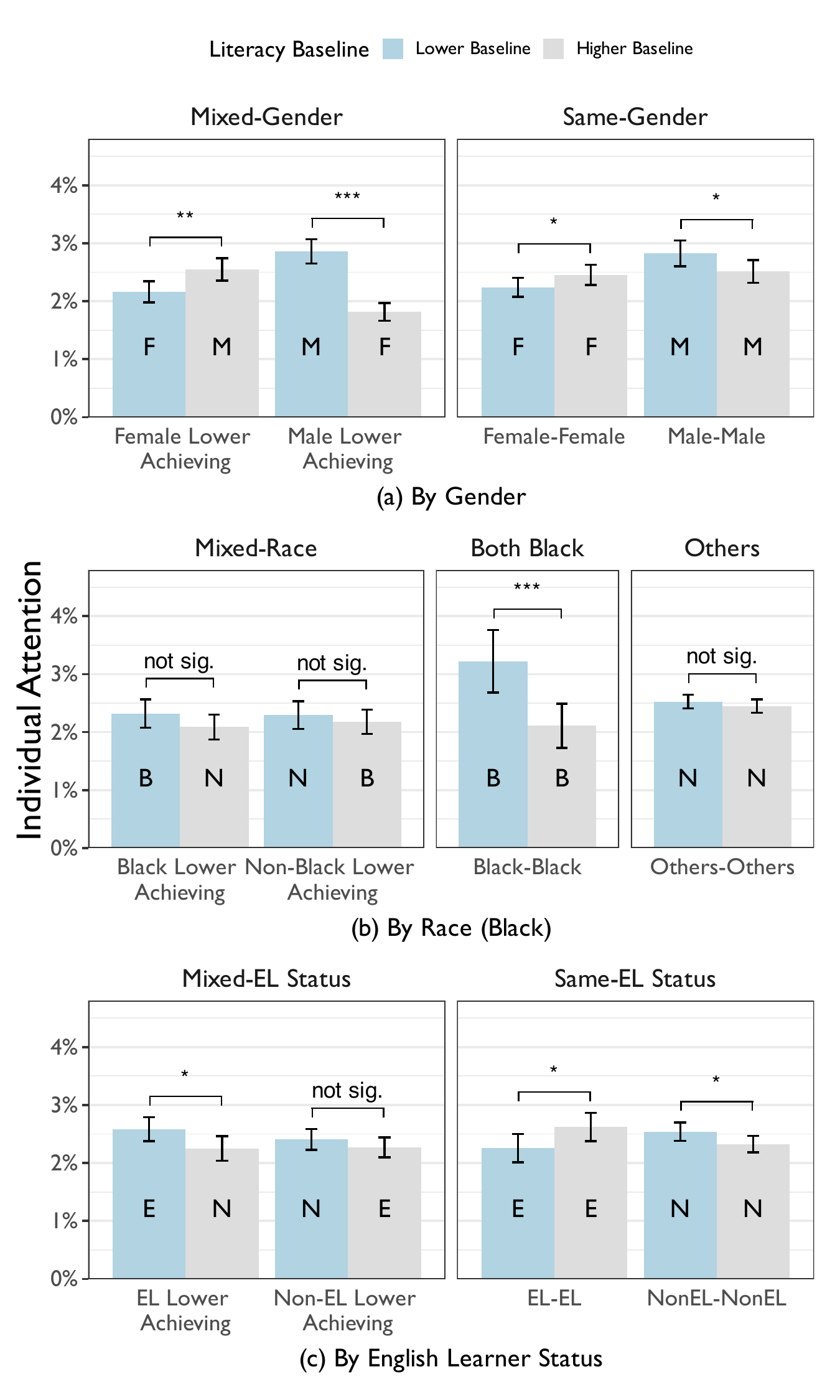}
        \caption{Management}
        \label{fig:study3_management}
    \end{subfigure}
    \caption{\textbf{Nature of attention by student achievement relative to their peer.}
    }
    \label{fig:study3}
\end{figure*}

The main study reports aggregate educator attention based on student demographics and pair composition.
Here, we break down these findings by the nature of attention (content, relationship-building, and management) reported in Figure~\ref{fig:study3}.

\paragraph{By Gender}
The main study reports that in mixed-gender pairs, lower-achieving female students not only fail to receive additional attention but receive significantly \textit{less} attention than their higher-achieving male peers.
Figure~\ref{fig:study3_management} reveals that this disparity is primarily driven by an increase in management-focused attention directed toward the male student.
In contrast, lower-achieving male students receive substantially more attention than their higher-achieving female peers, and this increase is observed across all attention categories. 

\paragraph{By Race}
The main study reports that in mixed-race pairs, lower-achieving Black students do not receive the additional attention we would expect.
They receive about the same amount of attention as their non-Black higher-achieving peer. 
However, when both students are Black, the lower-achieving Black student receives more attention than their higher-achieving Black peer, and this is driven by all forms of attention.

\paragraph{By EL Status}
The main study reports that in pairs with mixed EL status, the lower-achieving EL student receives more attention. 
This is primarily driven by attention on content and relationship-building. 
However, in pairs where both students have EL status, the trend of attending to the lower-achieving student reverses: 
the lower-achieving EL student receives \textit{less} attention than their higher-achieving EL peers, especially in content and relationship-building.

\subsection{Results on Attention for ``Both Students'' and ``One of the Students'' \label{app:add_framework_results}}

Our framework additionally identifies when educator utterance is directed to both students or directed to an ambiguous recipient.
We report those results in this section. 

\begin{table}[h] 
    \centering
        \begin{tabular}{lccc}
        \toprule
        \textbf{Pairings} & \textbf{Mean} & \textbf{Standard Deviation} & \textbf{n}\\  
        \midrule
        Mixed-Gender & $0.369$ & $0.160$ & $2543$\\
        Female-Female & $0.378$ & $0.157$ & $1606$\\
        Male-Male & $0.390$ & $0.157$ & $1100$\\
        \midrule
        Mixed-Race & $0.374$ & $0.157$ & $1619$\\
        Black-Black & $0.350$ & $0.146$ & $257$\\
        Others-Others & $0.379$ & $0.160$ & $3373$\\
        \midrule
        Mixed-EL Status & $0.383$ & $0.158$ & $2363$\\
        EL-EL & $0.403$ & $0.173$ & $741$\\
        NonEL-NonEL & $0.359$ & $0.151$ & $2101$\\
        \bottomrule
        \end{tabular}
        \vspace{1em}
    \caption{\textbf{Descriptive statistics on ``both students'' as the attention recipient.}
    }
    \label{tab:both}
\end{table}

\begin{table}[h] 
\centering    
    \begin{tabular}{@{\extracolsep{5pt}}lcc}  
    \toprule
    \textbf{Category} & \textbf{Pairings} & \textbf{Estimate} \\  
    \midrule
    By Gender  & Female-Female & $-0.015^{***}$ \\    
                 &    & $(0.004)$ \\    
               & Male-Male & $-0.042^{***}$ \\    
                 &    & $(0.005)$ \\ 
    \midrule
    By Race (Black)  & Black-Black & $0.034^{***}$ \\  
                 &    & $(0.009)$ \\    
                     & Others-Others & $-0.021^{***}$ \\    
                 &    & $(0.004)$ \\  
    \midrule
    By EL Status      & EL-EL & $0.011^{**}$ \\    
                 &    & $(0.006)$ \\    
                      & NonEL-NonEL & $0.001$ \\    
                 &    & $(0.004)$ \\    
    \midrule 
    Observations &    & $5,205$ \\  
    R$^{2}$      &    & $0.025$ \\  
    Adjusted R$^{2}$ & & $0.024$ \\  
    Residual Std. Error & & $0.129$ ($df = 5198$) \\  
    F Statistic  &    & $22.443^{***}$ ($df = 6; 5198$) \\  
    \midrule
    \textit{Note:}  & \multicolumn{2}{r}{$^{*}p<0.1; \quad ^{**}p<0.05; \quad ^{***}p<0.01$} \\
    \bottomrule
    \end{tabular} 
    \vspace{1em}
  \caption{\textbf{Regression-based check on ambiguous ``Only one'' recipient category.} 
  The heterogeneous pairing (e.g. mixed-gender) is set as the baseline.}
\label{tab:ambiguous}  
\end{table} 

Table \ref{tab:both} reports the amount of educator attention directed to both students across different pair compositions. 

The ambiguous ``One of the Students'' category captures utterances directed to only one student but the exact recipient is unclear.
One potential concern is whether the amount of ambiguous references is large enough to dispute our earlier findings. 
We conduct a regression-based check to test the robustness of our results. 
We model of the amount of ambiguous attention, $y_{k}$ in session k as function of indicators of their Gender pair composition $G_k$ (i.e. both female, both male, mixed-gender), Race (black) pair composition $R_k$, and EL status pair composition $E_k$.
\begin{equation*}
   y_{k} = \beta_{1}G_k + \beta_{2}R_k + \beta_{3}E_k + e_{k} 
\end{equation*}

Table~\ref{tab:ambiguous} reports the coefficients of the regression. 
We compare the coefficients against Study 2's results and confirm that our findings are robust.
Take gender pairings as an example. 
Table~\ref{tab:ambiguous} shows that compared to both-male sessions, mixed-gender sessions have 4.2 \pp more utterances classified as ambiguous. Meanwhile, we find in Study 2 that in mixed-gender pairings female students receive 5.2 \pp less attention than their male partners. 
Even if we assume that all of the additional 4.2 \pp of ambiguous utterances were actually directed to female students (but misclassified), females would still receive 1 \pp less attention than their male partners.
Thus, our earlier findings still hold.

\section{Additional Model Evaluations \label{app:add_model}}
To determine the best classification approach, we evaluated various models, methods and heuristics. 

We compare the following approaches in the table: 
\begin{itemize}
    \item Name in Text: A simple heuristic where the recipient is identified based solely on the presence of names in the target utterance text. If `[Student A]` appears in the utterance, then the utterance is classified as addressing Student A, same with `[Student B]` for Student B. If both names appear, then the utterance is classified as addressing both students. If no names appear, then the utterance is classified as addressing one of the students.
    \item Name in Text \& Context: A revised heuristic that follows the same rules as the previous heuristic, but check for names in the target utterance text and the preceding context (10 previous conversation turns).
    \item RoBERTa~\citep{liu2019robertaD}: Fine-tuned on the train split of our annotated dataset.
    \item GPT-4, Claude-Haiku, Claude-Sonnet, Claude-Opus: Closed-source models with varying performance and cost implications. We tested these models with a zero-shot classification setup or $k$-shot setup. Prompts for these setups are shown in Figures~\ref{fig:prompt0} (0-shot), \ref{fig:prompt1} (1-shot) and \ref{fig:prompt3} (3-shot).
\end{itemize}

We report the F1 score and the average API cost per 100 transcripts, based on pricing from OpenAI and Anthropic.
The performance metrics are reported in Table~\ref{tab:model_performance} on our original dataset for classifying recipient of attention. 
Given the cost-effectiveness and performance, we chose to finetune RoBERTa models for our study since they provided similar performance to the best closed-source options without incurring additional API expenses.

\begin{table}[h]
  \centering
  \small
  \begin{tabular}{cl|c|c|c}
    \toprule
    \multicolumn{1}{c}{\textbf{Method}} & \multicolumn{1}{c}{$\bf k$} & \multicolumn{1}{c}{\textbf{F1}} & \multicolumn{1}{c}{\textbf{Cost (\$)}} \\
    \midrule
    \multicolumn{1}{c}{\texttt{Name} in text} & \multicolumn{1}{c}{-} & \multicolumn{1}{c}{$0.279$} & \multicolumn{1}{c}{-} \\
    \multicolumn{1}{c}{\texttt{Name} in text \& context} & \multicolumn{1}{c}{-} & \multicolumn{1}{c}{$0.435$} & \multicolumn{1}{c}{-} \\
    \multicolumn{1}{c}{\texttt{RoBERTa}} & \multicolumn{1}{c}{-} & \multicolumn{1}{c}{\yellowcell $\bf 0.718$} & \multicolumn{1}{c}{-} \\
    \midrule
    \multicolumn{1}{c}{\texttt{GPT4}} & \multicolumn{1}{c}{0} & \multicolumn{1}{c}{$0.577$}  & \multicolumn{1}{c}{$62.1$} \\
    \multicolumn{1}{c}{\texttt{GPT4}} & \multicolumn{1}{c}{1} & \multicolumn{1}{c}{$0.607$}  & \multicolumn{1}{c}{ $83.4$} \\
    \multicolumn{1}{c}{\texttt{GPT4}} & \multicolumn{1}{c}{3} & \multicolumn{1}{c}{$0.654$}  & \multicolumn{1}{c}{ $ 128.0$} \\
    \midrule
    \multicolumn{1}{c}{\texttt{Claude-Haiku}} & \multicolumn{1}{c}{0} & \multicolumn{1}{c}{$0.514$}  & \multicolumn{1}{c}{ $1.82$} \\
    \multicolumn{1}{c}{\texttt{Claude-Haiku}} & \multicolumn{1}{c}{1} & \multicolumn{1}{c}{$0.443$}  & \multicolumn{1}{c}{ $2.37$} \\
    \multicolumn{1}{c}{\texttt{Claude-Haiku}} & \multicolumn{1}{c}{3} & \multicolumn{1}{c}{$0.530$ }  & \multicolumn{1}{c}{ $3.53$} \\
    \midrule
    \multicolumn{1}{c}{\texttt{Claude-Sonnet}} & \multicolumn{1}{c}{0} & \multicolumn{1}{c}{$0.629$}  & \multicolumn{1}{c}{ $21.83$} \\
    \multicolumn{1}{c}{\texttt{Claude-Sonnet}} & \multicolumn{1}{c}{1} & \multicolumn{1}{c}{$0.608$}  & \multicolumn{1}{c}{ $28.46$} \\
    \multicolumn{1}{c}{\texttt{Claude-Sonnet}} & \multicolumn{1}{c}{3} & \multicolumn{1}{c}{$ \faintedbluecell \bf 0.725$}  & \multicolumn{1}{c}{ $42.32$} \\
    \midrule
    \multicolumn{1}{c}{\texttt{Claude-Opus}} & \multicolumn{1}{c}{0} & \multicolumn{1}{c}{$0.655$}  & \multicolumn{1}{c}{$109.16$} \\
    \multicolumn{1}{c}{\texttt{Claude-Opus}} & \multicolumn{1}{c}{1} & \multicolumn{1}{c}{$0.670$}  & \multicolumn{1}{c}{$142.31$} \\
    \multicolumn{1}{c}{\texttt{Claude-Opus}} & \multicolumn{1}{c}{3} & \multicolumn{1}{c}{$0.698$}  & \multicolumn{1}{c}{ $211.59$} \\
    \bottomrule
  \end{tabular}
  \vspace{1em}
  \caption{
  \textbf{Classification performance.} Cost is reported as estimated cost for annotating 100 transcripts with the method.}
  \label{tab:model_performance}
\end{table} 

\begin{figure*}[t]
    \centering
    \small
    \begin{tcolorbox}[
    title={\textbf{Prompt for zero-shot classification}},
    ]
    Your task is to read the following conversation snippet and classify whom the tutor is talking to. The conversation comes from a K-2 early literacy tutoring session between a tutor and two students. These students have been de-identified as [Student A] and [Student B].\\
    
    The possible labels are:\\
    0: The tutor is addressing both students. e.g., "Let's do it together, [Student A], [Student B] and me."\\
    1: The tutor is addressing Student A. e.g., "Okay, [Student A], it's your turn."\\
    2. The tutor is addressing Student B. e.g., "Good job, [Student B]."\\
    3. The tutor is addressing one of the students, but it is unclear which one. e.g., "Let's wait for him."\\
    
    Only output the label number. Do not output anything else.\\
    
    Context: \{context\}\\
    Text: \{text\}\\
    Label (number):
    \end{tcolorbox}
    \caption{\textbf{Prompt for zero-shot classification used with GPT-4, Claude-Haiku, Claude-Sonnet, Claude-Opus.} \texttt{\{context\}} is the placeholder for the educator utterances (10 preceding lines) leading up to the target utterance for classification. \texttt{\{text\}} is the educator utterance we aim to assign label.
    \label{fig:prompt0}}
\end{figure*}

\begin{figure*}[t]
    \centering
    \small
    \begin{tcolorbox}[
    title={\textbf{Prompt for $k$-shot classification ($k=1$)}},
    ]
    Your task is to read the following conversation snippet and classify whom the tutor is talking to. The conversation comes from a K-2 early literacy tutoring session between a tutor and two students. These students have been de-identified as [Student A] and [Student B].\\
    
    The possible labels are:\\
    0: The tutor is addressing both students. e.g., "Let's do it together, [Student A], [Student B] and me."\\
    1: The tutor is addressing Student A. e.g., "Okay, [Student A], it's your turn."\\
    2. The tutor is addressing Student B. e.g., "Good job, [Student B]."\\
    3. The tutor is addressing one of the students, but it is unclear which one. e.g., "Let's wait for him."\\
    
    Only output the label number. Do not output anything else.\\
    
    Context: Don't do it, don't do it, don't do it, don't do it, don't do it. She can circle it, but this is, it's for [Student B] to circle. It's for [Student B] to circle. It's hers, because Gamela had a turn. We got to learn to take turns.\\
    Text: You got it, you have it, [Student B].\\
    Label (number): 2\\
    
    Context: \{context\}\\
    Text: \{text\}\\
    Label (number):
    \end{tcolorbox}
    \caption{\textbf{Prompt for $k$-shot classification ($k=1$) used with GPT-4, Claude-Haiku, Claude-Sonnet, Claude-Opus.} \texttt{\{context\}} is the placeholder for the educator utterances (10 preceding lines) leading up to the target utterance for classification. \texttt{\{text\}} is the educator utterance we aim to assign label.
    \label{fig:prompt1}}
\end{figure*}

\begin{figure*}[t]
    \centering
    \small
    \begin{tcolorbox}[
    title={\textbf{Prompt for $k$-shot classification ($k=3$)}},
    ]
    Your task is to read the following conversation snippet and classify whom the tutor is talking to. The conversation comes from a K-2 early literacy tutoring session between a tutor and two students. These students have been de-identified as [Student A] and [Student B].\\
    
    The possible labels are:\\
    0: The tutor is addressing both students. e.g., "Let's do it together, [Student A], [Student B] and me."\\
    1: The tutor is addressing Student A. e.g., "Okay, [Student A], it's your turn."\\
    2. The tutor is addressing Student B. e.g., "Good job, [Student B]."\\
    3. The tutor is addressing one of the students, but it is unclear which one. e.g., "Let's wait for him."\\
    
    Only output the label number. Do not output anything else.\\
    
    Context: Don't do it, don't do it, don't do it, don't do it, don't do it. She can circle it, but this is, it's for [Student B] to circle. It's for [Student B] to circle. It's hers, because Gamela had a turn. We got to learn to take turns.\\
    Text: You got it, you have it, [Student B].\\
    Label (number): 2\\
    
    Context: Who is like this or how? [Student A] has hair. How does she have hair? [Student A] has hair. have, however, horses have We missed it too. We can't- it has to start with the letter H. I'm gonna put headphones. This is funny. Oh [Student B], why do I keep doing that? There you go.\\
    Text: So that's- that's- that's the sentence.\\
    Label (number): 0\\
    
    Context: This just helps us kind of map out the sounds that we hear. Oh, your word is bonnet. I'm going to move them for you. No, your word is kitten, [Student B]. Are you missing any? Good job, [Student B]. OK. [Student A], I'm going to tell you your word one more time. Bonnet. So let's see where we can fix it.\\
    Text: Because you put bonnet.\\
    Label (number): 1\\
    
    Context: \{context\}\\
    Text: \{text\}\\
    Label (number):
    \end{tcolorbox}
    \caption{\textbf{Prompt for $k$-shot classification ($k=3$) used with GPT-4, Claude-Haiku, Claude-Sonnet, Claude-Opus.} \texttt{\{context\}} is the placeholder for the educator utterances (10 preceding lines) leading up to the target utterance for classification. \texttt{\{text\}} is the educator utterance we aim to assign label.
    \label{fig:prompt3}}
\end{figure*}

\end{document}

%% file: macros.tex
\newlength{\widebarargwidth}
\newlength{\widebarargheight}
\newlength{\widebarargdepth}

\newcommand\eg{e.g.,~}

\newcommand\pp{~p.p.~}
\newcommand\ppm{p.p.}

\newcommand\numUtterances{~1,157,970~}
\newcommand\numTranscripts{~5,249~}
\newcommand\numAnnotations{~1,200~}
\newcommand\numTeachers{~125~}
\newcommand\numStudents{~793~}

%% file: std_macros.tex
\ifthenelse{\isundefined{\definition}}{\newtheorem{definition}{Definition}}{}
\ifthenelse{\isundefined{\assumption}}{\newtheorem{assumption}{Assumption}}{}
\ifthenelse{\isundefined{\hypothesis}}{\newtheorem{hypothesis}{Hypothesis}}{}
\ifthenelse{\isundefined{\proposition}}{\newtheorem{proposition}{Proposition}}{}
\ifthenelse{\isundefined{\theorem}}{\newtheorem{theorem}{Theorem}}{}
\ifthenelse{\isundefined{\lemma}}{\newtheorem{lemma}{Lemma}}{}
\ifthenelse{\isundefined{\corollary}}{\newtheorem{corollary}{Corollary}}{}
\ifthenelse{\isundefined{\alg}}{\newtheorem{alg}{Algorithm}}{}
\ifthenelse{\isundefined{\example}}{\newtheorem{example}{Example}}{}

%% file: images/framework.tex
\begin{table*}[t]
\normalsize
\centering
\renewcommand{\arraystretch}{1.6} 
\setlength{\tabcolsep}{6pt} 
\resizebox{\textwidth}{!}{%
\begin{tabular}{cc|p{0.3\textwidth} p{0.3\textwidth} p{0.3\textwidth}}
\toprule
\multicolumn{2}{c}{} & \multicolumn{3}{c}{\textbf{\Large\textsc{Nature of Attention}}} \\ 
\multicolumn{2}{c}{} & \multicolumn{1}{c}{\large\textbf{Content}} & \multicolumn{1}{c}{\large\textbf{Relationship-Building}} & \multicolumn{1}{c}{\large\textbf{Management}} \\ 
\midrule
\multirow{4}{*}{\rotatebox[origin=c]{90}{\Large\textbf{\textsc{\parbox[c]{5cm}{\centering Recipient of \\ Attention}}}}} 
& \large\textbf{Only Student A} & "Okay, [Student A], can you use this word in a sentence, please?" & "Okay, [Student A], tell me one fun thing from the weekend." & "Oh, [Student A], I can't see you." \\ 
& \large\textbf{Only Student B} & "Okay, [Student B], what's the middle sound?" & "I know [Student B] likes to read." & "And [Student B], can you hit mute?" \\ 
& \large\textbf{Both Students} & "Ooh, mushroom does have the M sound." & "You guys are awesome." & "Let's keep our listening ears on and our focus, and let's do this page." \\ 
& \large\textbf{One of the Students} & "Ball." & "Today you're six years old?" & "Okay, keep your earphones on, stop moving the computer screen." \\ 
\bottomrule
\end{tabular}
}
\caption{
\textbf{A sample of utterances illustrating our two-dimensional Attention Framework.} 
Our framework classifies a teacher utterance based on both the \textsc{Nature of Attention} (content, relationship-building, or management) and the \textsc{Recipient of Attention}. 
The classification of each utterance relies on the broader conversational context (not shown in the table). 
While some utterances can be clearly categorized, many require careful consideration of context for accurate classification. 
\label{tab:framework}}
\end{table*}